%% file: main.tex
\documentclass{article}

\PassOptionsToPackage{numbers,compress}{natbib}
\usepackage[preprint]{neurips_2026}

\usepackage[utf8]{inputenc}
\usepackage[T1]{fontenc}
\usepackage{hyperref}
\usepackage{url}
\usepackage{amsmath,amssymb,amsfonts}
\usepackage{algorithmic}
\usepackage{graphicx}
\usepackage{textcomp}
\usepackage{xcolor}
\usepackage{multirow}
\usepackage{booktabs}
\usepackage{svg}
\usepackage{enumitem}
\usepackage{makecell}  
\usepackage{algorithm}
\usepackage{subcaption}
\usepackage{wrapfig}

\begin{document}
\providecommand{\nolinenumbers}{}
\nolinenumbers

\title{LLMForge: Multi-Backend Hardware-Aware Neural Architecture Search with Infinite-Head Attention for Edge Language Models}

\author{%
  Xinting Jiang$^{1}$, Junyi Luo$^{1}$, Ruichen Qi$^{1}$ \\
  \bfseries Kauna Lei$^{2}$, Ben Laurie$^{3}$, Gregory Kielian$^{3}$, Mehdi Saligane$^{1}$ \\[0.4em]
  $^{1}$Brown University \quad
  $^{2}$University of Michigan \quad
  $^{3}$Google Research
}

\maketitle
\input{sections/0-abstract}

\input{sections/1-introduction}
\input{sections/2-related_work}
\input{sections/3-method}

\input{sections/4-result_and_analysis}
\input{sections/5-conclusion}

\clearpage
\bibliographystyle{plainnat}
\bibliography{ref}

\input{sections/7-appendix}

\end{document}

%% file: sections/0-abstract.tex
\begin{abstract}
Sub-billion-parameter Transformer language models are increasingly deployed on edge devices, where the privacy, latency, and operating-cost advantages of on-device inference are constrained by tight memory-bandwidth, energy, and thermal budgets that make architectural choice and accelerator-specific cost central to efficient inference.
We present \textbf{LLMForge}, a hardware-aware neural architecture search (NAS) framework whose three composable contributions together make edge-LM architecture search hardware-conditioned, since different substrates impose different hardware cost bottlenecks.
\textbf{Infinite-Head Attention (IHA)} decouples the number of query heads, KV groups, and per-head query/key and value dimensions, expanding the feasible per-layer attention configuration space by $\sim\!400\times$ over grouped-query attention within our search-space ranges.
\textbf{Forge-Former}, an encoder-based surrogate for ranking architectural candidates, outperforms MLP and random-forest baselines.
\textbf{Forge-DSE}, an NSGA-II-based design-space-exploration engine, pairs Forge-Former with a multi-backend hardware cost model spanning GPUs, systolic accelerators, and ring-dataflow edge accelerators.
Across four different hardware substrates, the searches converge to visibly different architectures whose shapes track each substrate's cost bottleneck.
On the multi-chip ring substrate, our co-search returns three 300M-scale deployment-aware variants on the Pareto front.
Each is re-trained on FineWeb-Edu-10BT under matched recipe against SmolLM2-360M and Qwen-0.5B architecture baselines.
The accurate variant has the lowest validation loss $2.798$ and competitive benchmark performance with fewer parameters, the energy-optimized variant lowers energy per token by $40\%$, and the latency-optimized variant lowers TTFT and TPOT by $43\%$.
\end{abstract}

%% file: sections/1-introduction.tex
\section{Introduction}
\label{sec:introduction}

Transformer-based language models~\citep{vaswani2017attention} now power a wide range of edge workloads,
from on-device translation~\citep{parry2021dynamictransformerefficientmachine}
to real-time speech recognition~\citep{conformer-based-speech,latif2023transformers}.
On-device deployment is attractive for privacy, reliability, latency, and operating cost,
yet recent mobile-inference studies show that on-device LLM execution is memory-bound and energy-limited under tight thermal budgets~\citep{laskaridis2024meltingpoint},
a regime in which autoregressive decoding repeatedly incurs expensive weight and KV-cache memory traffic.
Post-training compression techniques such as quantization and pruning reduce these costs incrementally
but leave the underlying architectural degrees of freedom unexplored.

What architectural choices lead to edge efficiency under restrained parameter and resource budgets remains an open question.
Heuristic studies have surfaced layer-shape design rules at sub-billion scale, but a systematic study of the joint design space is still limited.
Hardware-aware neural architecture search (NAS) for Transformers, surveyed in \S\ref{sec:related}, offers a natural path, yet existing pipelines tend to leave three pieces of the search loop loose.
First, the attention parameterizations commonly carried over from prior NAS work fix the per-head dimension and tie key/value capacity, leaving the per-layer attention shape mostly determined by a single head-count choice.
Second, the evaluation backend is the search-cost bottleneck, since training each candidate from scratch is computationally expensive.
Third, hardware cost is often modeled as a single accelerator class or through coarse proxies such as parameter count or KV-cache size, which can mask the substrate-dependent re-ranking of candidates.

We present \textbf{LLMForge}, a hardware-aware NAS framework whose three composable contributions plug into a single NSGA-II~\citep{nsgaII} loop.

\paragraph{1. Infinite-Head Attention (IHA).}
IHA treats the attention shape parameters $n_h$, $n_{kv}$, $d_{qk}$, $d_v$ as independent per-layer variables,
removing the divisibility and Q/K--V coupling constraints of multi-head attention.
Under the ranges in Table~\ref{tab:search_space},
this expands the feasible attention configuration space from $27$ under grouped-query attention to $11{,}250$ under IHA, a $\sim\!400\times$ per-layer expansion,
as detailed in \S\ref{subsec:iha}.

\paragraph{2. Forge-Former.}
Forge-Former is a Transformer-encoder accuracy surrogate that reaches Spearman $\rho = 0.75$ and Kendall $\tau = 0.58$ on the IHA test split, $1.4$ to $2.5\times$ above MLP and random-forest baselines,
enabling NSGA-II selection at a fraction of per-candidate train-from-scratch cost,
as detailed in \S\ref{subsec:forge_former}.

\paragraph{3. Forge-DSE.}
Forge-DSE is an NSGA-II design-space-exploration engine that pairs Forge-Former with a multi-backend cost model across GPUs, systolic accelerators, and ring-dataflow edge accelerators,
and can optionally co-evolve the surrogate during search, as detailed in \S\ref{subsec:forge_dse}.
Across four different hardware substrates, Forge-DSE produces substrate-conditioned Pareto fronts that reshape per-layer width and depth to each substrate's cost bottleneck.
On the rDXE substrate, joint architecture-and-chip co-search returns a deployment-aware family that lowers validation loss, energy per token, and latency below the SmolLM2-360M baseline at fewer parameters.

%% file: sections/2-related_work.tex
\section{Related Work}
\label{sec:related}

\textbf{Heterogeneous edge Transformer architectures.}
Reducing inference cost at edge scale has motivated design space exploration of edge-scale language model architectures.
Within standard attention, multi-query attention (MQA)~\citep{shazeer2019mqa} shares a single K/V projection across query heads, and grouped-query attention (GQA)~\citep{ainslie2023gqa} interpolates between MQA and multi-head attention (MHA) through $n_{kv}$ K/V groups.
Multi-head latent attention (MLA)~\citep{deepseekv2} instead compresses K/V through a shared low-rank latent, decoupling key/value capacity from the per-head dimension but fixing the latent dimension as a global hyperparameter.
Beyond standard attention, Falcon-H1~\citep{zuo2025falconh1} couples Transformer attention with state-space mixers in a parallel hybrid head, and Jet-Nemotron~\citep{gu2025jetnemotron} introduces JetBlock, a linear-attention block with input-conditioned dynamic convolutions and gating, both replacing the attention primitive itself.
Composer~\citep{acun2025composer} keeps standard attention and MLP primitives but searches over their per-layer interleaving via Bayesian optimization, discovering non-uniform attention-to-MLP ratios.
At the layer-shape level, MobileLLM~\citep{liu2024mobilellm} and OpenELM~\citep{mehta2024openelm} validate that thinner-deeper backbones and progressively allocated widths matter at sub-billion scale.
Infinite-Head Attention (IHA) relaxes these MHA constraints, broadening the per-layer attention design space. Our framework pairs IHA with joint search over per-layer parameterization and layer depth.


\textbf{Hardware-aware NAS.}
Hardware-aware NAS for Transformers has been pursued via a range of search methods and hardware-cost models.
HAT~\citep{wang2020hat} pairs an evolutionary search over a weight-sharing supernet with per-device latency lookup tables to find Transformer architectures with low device-specific inference latency.
TransCODE~\citep{transcode} co-optimizes Transformer architecture and ASIC accelerator design via Bayesian active learning, treating the substrate as part of the search rather than a deployment target.
STAR~\citep{thomas2024star} runs an evolutionary search over a hybrid genome but proxies hardware cost with parameter count and KV-cache size in place of analytical cost models.
HW-GPT-Bench~\citep{sukthanker2024hwgptbench} releases a benchmark of per-device latency and energy surrogates for a fixed GPT-2-style decoder-only architecture space across multiple GPU and CPU targets.
LLMForge differs from this line by pairing a broadened per-layer attention search space with a multi-backend cost model and an encoder-based accuracy surrogate, producing substrate-conditioned Pareto fronts rather than a single architecture per search.

%% file: sections/3-method.tex
\section{Method}
\label{sec:method}

\begin{figure*}[htbp]
\centering
\includegraphics[width=\textwidth]{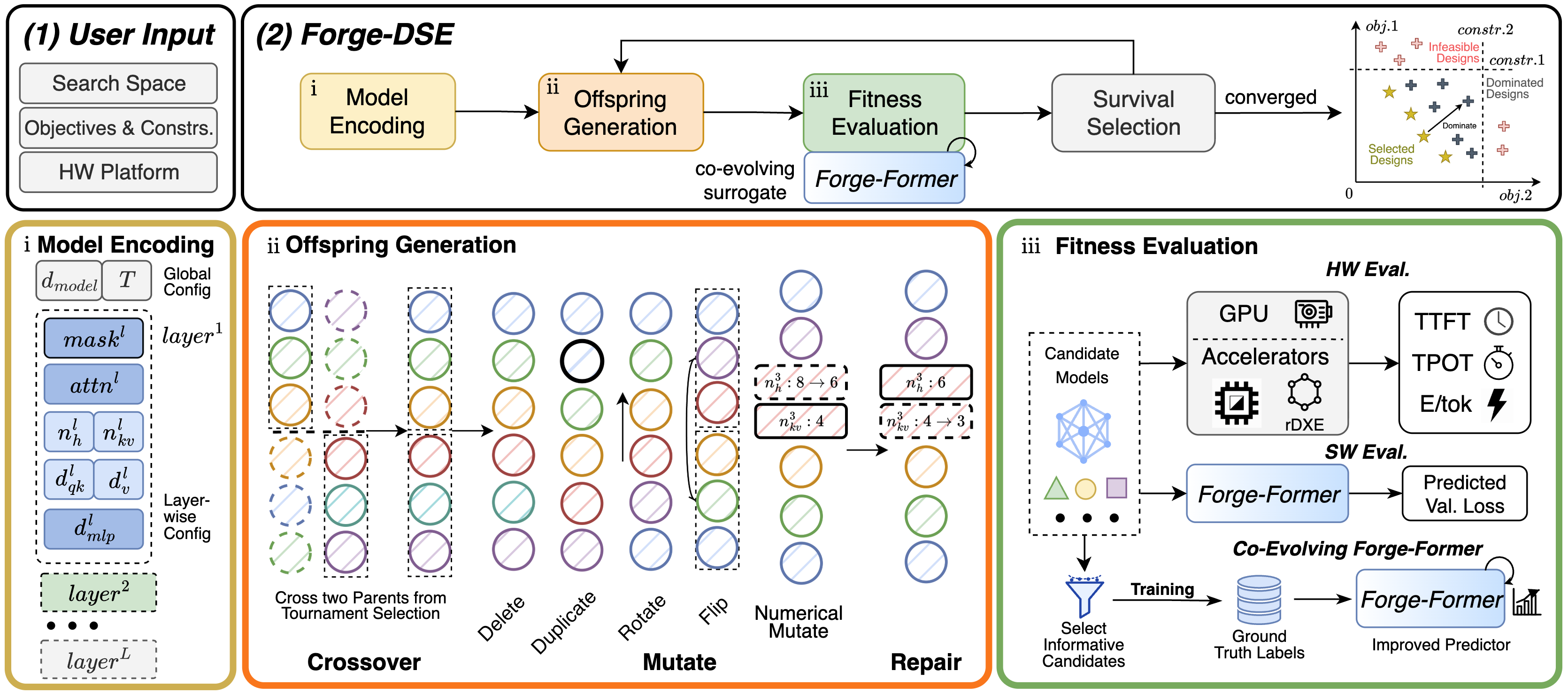}
\caption{Forge-DSE pipeline. Top: the four-stage outer loop with co-evolving Forge-Former feedback. Bottom: zoom of the first three stages, (i) encoding, (ii) offspring generation, and (iii) fitness evaluation.}
\label{fig:framework_overview}
\end{figure*}

LLMForge runs a multi-objective evolutionary search over IHA-parameterized Transformer architectures.
Forge-Former predicts each candidate's validation loss at a fraction of train-from-scratch cost, a pluggable hardware backend reports deployment cost, and an optional surrogate-refinement event keeps Forge-Former aligned with the evolving population.
Figure~\ref{fig:framework_overview} summarizes this pipeline.
We detail each in turn: IHA in \S\ref{subsec:iha}, Forge-Former in \S\ref{subsec:forge_former}, and Forge-DSE in \S\ref{subsec:forge_dse}.

\begin{figure}[!ht]
\centering
\includegraphics[width=0.8\textwidth]{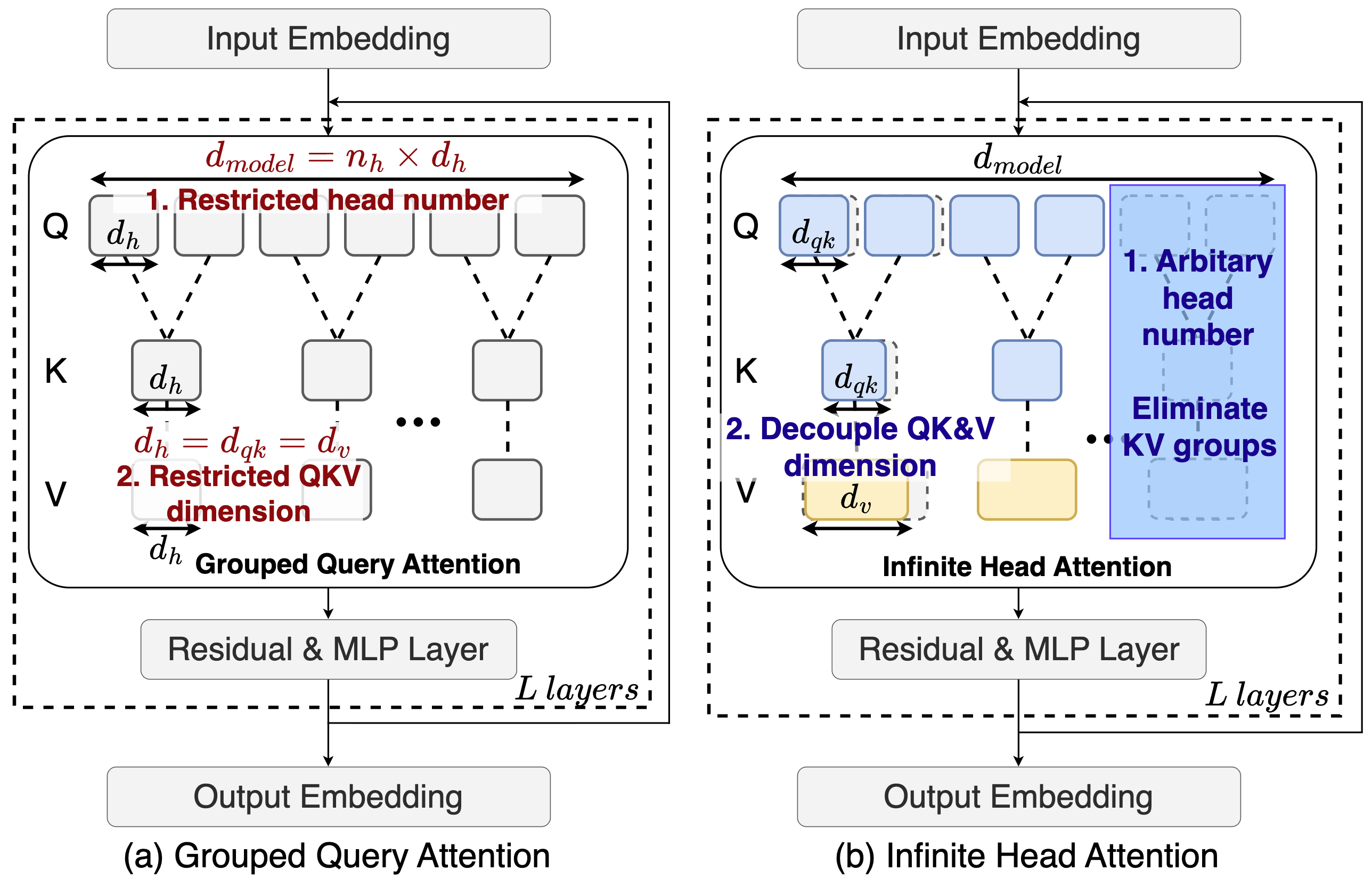}
\caption{Infinite-Head Attention (IHA). $n_h$, $n_{kv}$, $d_{qk}$, $d_v$ vary independently per layer; head outputs are concatenated and projected back to $d_{\text{model}}$.}
\label{fig:iha}
\end{figure}

\subsection{Infinite-Head Attention (IHA)}
\label{subsec:iha}

Multi-head attention couples attention-related shape parameters through two constraints: the \emph{divisibility constraint} $d_{\text{model}} = n_h \cdot d_h$ and the \emph{Q/K--V coupling} $d_h = d_{qk} = d_v$.
Under tight parameter budgets, these constraints are restrictive: increasing the head count forces a smaller per-head capacity, entangling head count with expressivity per head, and tying $d_v$ to $d_{qk}$ couples value-side and query/key-side capacity even though the output projection back to $d_{\text{model}}$ does not require it.

IHA relaxes both MHA constraints by treating $n_h$, $n_{kv}$, $d_{qk}$, and $d_v$ as independent per-layer parameters, as shown in Figure~\ref{fig:iha}. The only alignment requirement it preserves is the GQA grouping condition $n_{kv}\,|\,n_h$.
Attention is computed over $n_{kv}$ K/V groups, each serving $R = n_h / n_{kv}$ query heads under the standard grouped-query mapping $g\colon[1,n_h]\to[1,n_{kv}]$ given by $g(h) = 1 + \lfloor(h{-}1)/R\rfloor$~\citep{ainslie2023gqa}.

Decoupling these four shape parameters substantially expands the per-layer attention configuration count.
Under the example per-layer ranges of Appendix~\ref{app:search_space},
the feasible attention configuration count per layer grows from $27$ under the GQA divisibility and coupling constraints to $11{,}250$ under IHA, an expansion of $\sim\!400\times$.

\subsection{Forge-Former: Encoder-Based Architecture Surrogate}
\label{subsec:forge_former}

\begin{wrapfigure}[26]{R}{0.45\textwidth}
\centering
\vspace{-0.6em}
\includegraphics[width=0.43\textwidth]{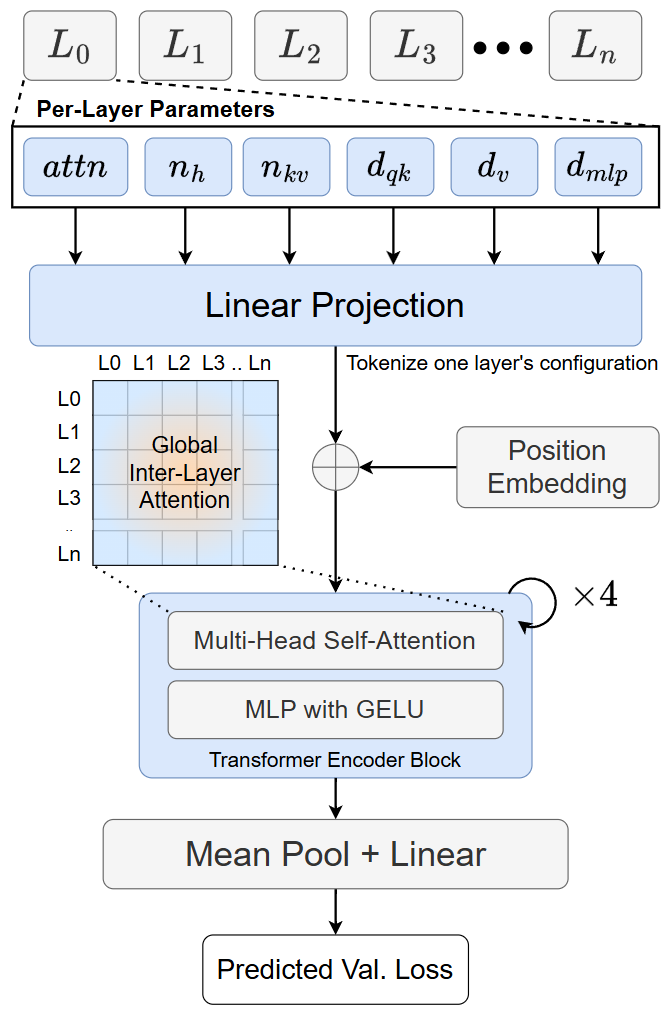}
\caption{Forge-Former architecture}
\label{fig:forge_former}
\vspace{-0.6em}
\end{wrapfigure}

Forge-Former is a learned surrogate $\hat{y}\colon \mathcal{A} \to \mathbb{R}_{>0}$ that maps an IHA-parameterized architecture $x \in \mathcal{A}$ to its predicted validation loss.
Figure~\ref{fig:forge_former} illustrates the Forge-Former architecture. Each architecture $x$ is encoded as a sequence of $L$ layer tokens, where each token concatenates the layer's IHA parameters.
Each scalar field is lifted to Forge-Former's model dimension by its own linear layer, and the per-layer token embedding is the sum of these per-field vectors plus a learned positional embedding over layer index.
Variable depth is handled by padding to $L$ with a per-sequence mask that excludes inactive layers from attention and from the pooled representation.
A pre-LN Transformer encoder over the layer-token sequence captures within-layer field interactions through its feedforward sublayers and cross-layer interactions through self-attention, and a masked mean over active tokens feeds a linear regression head that outputs $\hat{y}(x)$.

We train Forge-Former end-to-end with an $L_1$ regression objective on over 2k random sampled IHA architectures labeled with their validation loss under the protocol of Appendix~\ref{app:sw_eval}.
Ranking and regression metrics on the held-out test split are reported in \S\ref{sec:results}, and encoder hyperparameters and the optimizer are listed in Appendix~\ref{app:forge_former}.

\subsection{Forge-DSE: Multi-Backend Design-Space-Exploration Engine}
\label{subsec:forge_dse}

Forge-DSE is the multi-objective design-space-exploration engine of LLMForge.
It runs an evolutionary search loop over IHA-parameterized architectures, with software evaluation by Forge-Former or by training each candidate from scratch, paired with hardware cost from a selected backend.
Using Forge-Former restricts the search to the standard IHA design space of Appendix~\ref{app:search_space} that its training corpus covers, while training from scratch supports any IHA design space, including ones shaped around a seed architecture for tight-constraint backends such as the rDXE ring.
\S\ref{subsec:forge_dse:loop} describes the search loop and the optional surrogate co-evolution.
\S\ref{subsec:forge_dse:backends} describes the three hardware cost backends.

\subsubsection{Search Loop with Co-Evolving Forge-Former}
\label{subsec:forge_dse:loop}

Forge-DSE evolves populations of IHA-parameterized architectures via custom mutation operators paired with NSGA-II--style non-dominated sorting and crowding-distance survival~\citep{nsgaII}, under the procedure summarized in Figure~\ref{fig:framework_overview}, with full pseudocode in Algorithm~\ref{alg:forge_dse} of Appendix~\ref{app:forge_dse_alg}.
Each candidate is encoded as a global configuration plus per-layer settings. Parents are selected by binary tournament, recombined via single-point crossover, and mutated via four operators applied sequentially: \emph{deletion} (flips a layer's gating mask), \emph{duplication} (copies a preceding layer to reuse partially optimized structure), \emph{rotation/reflection} (permutes layer order), and \emph{numerical perturbation} of per-layer parameters.
The feasibility repair step of \S\ref{subsec:iha} is applied to each offspring.
If Forge-Former is used, each offspring is evaluated by Forge-Former for predicted validation loss $\hat{y}(x)$ with MC-dropout uncertainty $\sigma(x)$ from $n_{\mathrm{mc}}$ stochastic forward passes,
and by the selected hardware backend for hardware cost.
The combined parent--offspring pool undergoes non-dominated-sorting survival with feasibility preference and crowding-distance diversity.

When co-evolution is enabled, every $K$ generations the loop fires a \emph{surrogate-refinement event} that keeps Forge-Former aligned with the evolving population.
The event picks informative samples from the current population, real-trains them under the protocol of Appendix~\ref{app:sw_eval},
and fine-tunes Forge-Former from its frozen baseline checkpoint on the accumulated label buffer blended with the original training corpus as experience replay.
This co-evolution is intended to correct distribution shift as the search concentrates on regions the static training corpus undersamples.

\subsubsection{Hardware Cost Backends}
\label{subsec:forge_dse:backends}

Each backend implements the interface $\mathrm{HW}(x, \mathcal{W}) \to (\text{energy per token}, \mathrm{TTFT}, \mathrm{TPOT})$, mapping an IHA architecture $x$ under a fixed prefill$+$decode workload $\mathcal{W}$ to token-level cost.
Forge-DSE provides three backends: a measured GPU (Backend A), Timeloop-modeled substrates (Backend B), and a multi-chip ring-dataflow simulator (Backend C).
Backends B and C are analytical and share a per-operator pipeline: a workload analyzer decomposes each layer into primitive operators, and a backend-specific mapper returns per-operator latency and energy that are aggregated to token-level metrics.

\textbf{Backend A: ZEUS-measured GPU cost.} On a local NVIDIA A100, ZEUS~\citep{you2023zeus} measures energy per token, TTFT, and TPOT end-to-end under $\mathcal{W}$, supplying measured cost as the GPU-substrate reference.

\textbf{Backend B: Timeloop-modeled substrates.} Timeloop~\citep{timeloop} maps each primitive operator onto a target accelerator specification. We instantiate four substrates: a Gemmini-style systolic array~\citep{genc2021gemmini}, an Eyeriss-style row-stationary dataflow~\citep{chen2019eyerissv2}, a FLAT-style fused-attention dataflow~\citep{kao2023flat}
and a single-chip decoder-execution (DXE) accelerator~\citep{tao2026rdxe} which serves as the per-chip building block of the ring backend below.

\begin{figure*}[t]
\centering
\includegraphics[width=\textwidth]{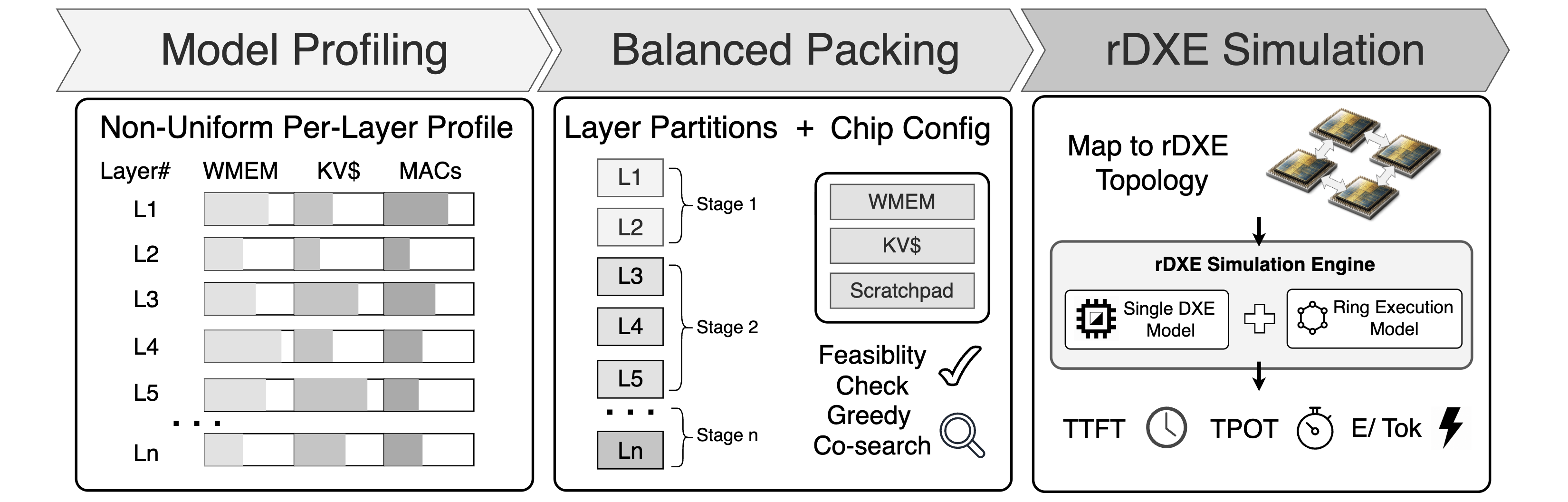}
\caption{Ring-dataflow co-search pipeline used by Backend C. \textit{Left:} per-layer resource profile (WMEM, KV\$, MACs). \textit{Middle:} multi-resource balanced packing finds a partition and shared DXE configuration. \textit{Right:} rDXE simulation engine (ring-execution $+$ single-DXE models) returns TTFT, TPOT, and energy per token.}
\label{fig:ring_co_search}
\end{figure*}

\textbf{Backend C: Ring-dataflow edge accelerator.} Backend C realizes the rDXE design of concurrent silicon-validated work~\citep{tao2026rdxe}, extending the single-chip DXE substrate of Backend B to a ring of multiple DXE chips that pipeline tokens across stages with stationary weights. The backend exposes a three-stage co-search pipeline (Figure~\ref{fig:ring_co_search}). (i) Each IHA layer is profiled for weight-memory, on-chip KV-cache, and MAC count. (ii) For each chip configuration in the swept chip-template grid of Appendix~\ref{app:ring_substrate}, a balanced packing routine assigns layers contiguously to ring stages, combining the greedy contiguous partition algorithm described in Appendix~\ref{alg:greedy_partition} with a binary search over a scalar per-stage decode-ops budget that minimises the longest-stage decode latency. (iii) An rDXE simulation engine composes a ring-execution and a single-DXE model to return energy per token, TTFT, and TPOT. All three metrics serve as Backend-C search objectives and are reported for Pareto-front architectures in \S\ref{sec:results}. The full substrate specification is detailed in~\citet{tao2026rdxe}.

%% file: sections/4-result_and_analysis.tex
\section{Results and Analysis}
\label{sec:results}

We conducted two sets of LLMForge experiments.
First, Forge-Former drives the search across four hardware substrates, reported in \S\ref{subsec:main_search}.
Second, every candidate is trained from scratch and the search co-optimises architecture together with the rDXE ring chip configuration, reported in \S\ref{subsec:scaled_validation}.
One is seeded from SmolLM2-135M for the $\sim$100M tier and the other one is from SmolLM2-360M for the $\sim$300M tier.
Forge-Former validation appears in \S\ref{subsec:forge_former_ablation}, with a search-recipe ablation in \S\ref{subsec:main_search}.

\subsection{Forge-Former Validation}
\label{subsec:forge_former_ablation}

We validate Forge-Former on HW-GPT-Bench's released dataset~\citep{sukthanker2024hwgptbench}, which tests the recipe at parity with their Net surrogate on a fixed decoder-only space, and on our IHA dataset (over 2k pairs labeled under Appendix~\ref{app:sw_eval}).
Primary metrics are ranking fidelity (Kendall-$\tau$, Spearman-$\rho$) and top-$X$\% recovery ($k$@$X$\%); MAE is reported for completeness.

\begin{table}[t]
\centering
\small
\caption{Surrogate accuracy on held-out test splits, mean\,$\pm$\,std across 5 random seeds. \emph{MLP} is HW-GPT-Bench's Net baseline~\citep{sukthanker2024hwgptbench} (one-hot encoded on \texttt{gpt\_l}, retrained on IHA fields); \emph{RF} is a random forest fit on the same train rows. MAE@5\% is mean absolute error restricted to the true top-5\% of test (Pareto fidelity); $k$@$X$\% is the smallest $k$ such that the predicted top-$k$ contains all true-top-$X$\%. Best per row in \textbf{bold}.}
\label{tab:predictor_accuracy}
\setlength{\tabcolsep}{4pt}
\begin{tabular}{@{}lccc@{\hspace{1em}}cc@{}}
\toprule
                & \multicolumn{3}{c}{IHA ($n_{\text{test}}{=}411$)} & \multicolumn{2}{c}{HW-GPT-Bench \texttt{gpt\_l} ($n_{\text{test}}{=}3000$)} \\
\cmidrule(lr){2-4} \cmidrule(lr){5-6}
                & \textbf{Forge-Former} & \textbf{MLP} & \textbf{RF} & \textbf{Forge-Former} & \textbf{MLP} \\
\midrule
MAE             & $\mathbf{0.213\,\pm\,0.011}$ & $0.483\,\pm\,0.025$ & $0.424\,\pm\,0.024$ & $0.064\,\pm\,0.007$ & $\mathbf{0.042\,\pm\,0.025}$ \\
MAE@5\%         & $\mathbf{0.057\,\pm\,0.020}$ & $0.268\,\pm\,0.044$ & $0.255\,\pm\,0.019$ & $\mathbf{0.027\,\pm\,0.015}$ & $0.038\,\pm\,0.018$ \\
Spearman $\rho$ & $\mathbf{0.754\,\pm\,0.022}$ & $0.345\,\pm\,0.019$ & $0.538\,\pm\,0.026$ & $0.998\,\pm\,0.001$ & $\mathbf{0.999\,\pm\,0.000}$ \\
Kendall $\tau$  & $\mathbf{0.582\,\pm\,0.016}$ & $0.235\,\pm\,0.015$ & $0.378\,\pm\,0.021$ & $0.966\,\pm\,0.010$ & $\mathbf{0.980\,\pm\,0.004}$ \\
$k$@1\%         & $\mathbf{96.8\,\pm\,32.0}$ & $198.2\,\pm\,86.3$ & $131.8\,\pm\,23.4$ & $54.4\,\pm\,30.0$ & $\mathbf{50.2\,\pm\,11.8}$ \\
$k$@5\%         & $\mathbf{261.8\,\pm\,80.2}$ & $360.6\,\pm\,41.2$ & $348.4\,\pm\,50.1$ & $\mathbf{221.8\,\pm\,27.8}$ & $302.2\,\pm\,86.4$ \\
\bottomrule
\end{tabular}
\end{table}

\begin{figure}[t]
\centering
\includegraphics[width=\textwidth]{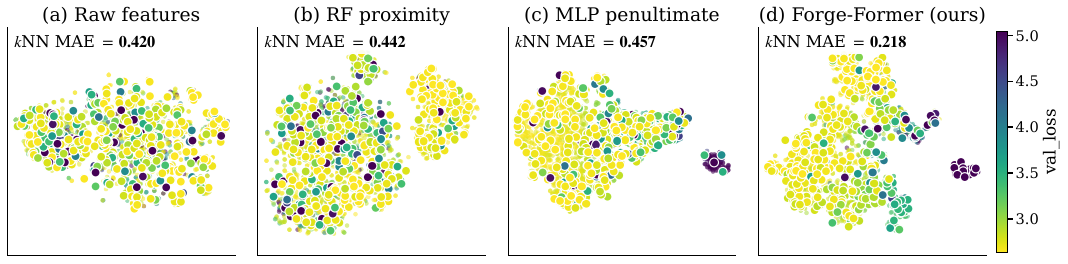}
\caption{t-SNE projections of held-out architecture embeddings, colored by validation loss. Each panel reports $k$NN val-loss MAE; lower values indicate that geometric proximity better reflects performance similarity.}
\label{fig:tsne_embedding}
\end{figure}

On HW-GPT-Bench's \texttt{gpt\_l} split, Forge-Former ties the Net baseline on global rank correlation while leading on MAE@5\% and $k$@5\%.
On the IHA dataset, Forge-Former leads on every metric, reaching Spearman $\rho = 0.75$ and Kendall $\tau = 0.58$, and lowering MAE@5\% by $\sim 5\times$ relative to both baselines (Table~\ref{tab:predictor_accuracy}).
The encoder's inter-layer self-attention gives Forge-Former an inductive bias for cross-layer interactions that a flat MLP on concatenated fields lacks. 
In the t-SNE analysis shown in Figure~\ref{fig:tsne_embedding}, the Forge-Former yields the lowest kNN MAE of $0.218$ and pulls the high-loss tail into a tight cluster rather than diffusing it through the embedding.

\subsection{Surrogate-Driven Multi-Backend Search}
\label{subsec:main_search}

This subsection covers the four Forge-Former-driven searches over the standard IHA design space.
Each search runs NSGA-II with Forge-Former and periodic surrogate refinement on one substrate from \S\ref{subsec:forge_dse:backends}, namely the ZEUS-measured A100 GPU and the three Timeloop-modeled systolic substrates Gemmini, Eyeriss, and FLAT.

\begin{figure}[t]
\centering
\includegraphics[width=\textwidth]{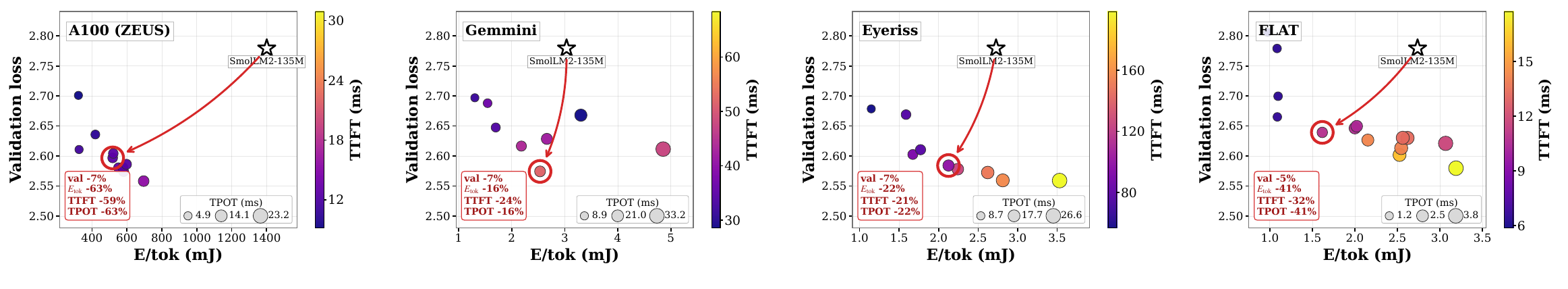}
\caption{Pareto fronts on the four Forge-Former-driven substrates. Each panel shows the architectures that lie on the Pareto front of the four search objectives. Marker colour encodes TTFT, marker size encodes TPOT, and the open star marks the SmolLM2-135M baseline measured on each substrate. The full search scatter plots, including all generations and the surrogate co-evolution trajectories, are shown in Appendix~\ref{app:cosearch_full_scatter}.}
\label{fig:substrate_pareto_4row}
\end{figure}

Figure~\ref{fig:substrate_pareto_4row} highlights one Pareto-front pick per substrate that strictly dominates the SmolLM2-135M anchor on all four objectives.
Across the four substrates, reductions vs the anchor range from $5\%$ to $7\%$ on validation loss, $16\%$ to $63\%$ on energy per token, $21\%$ to $59\%$ on TTFT, and $16\%$ to $63\%$ on TPOT.
Per-substrate breakdowns appear in Appendix~\ref{app:cosearch_full_scatter} and per-layer pick configurations in Appendix~\ref{app:pareto_archs}.

\begin{wrapfigure}[19]{R}{0.43\textwidth}
\centering
\vspace{-0.6em}
\includegraphics[width=0.43\textwidth]{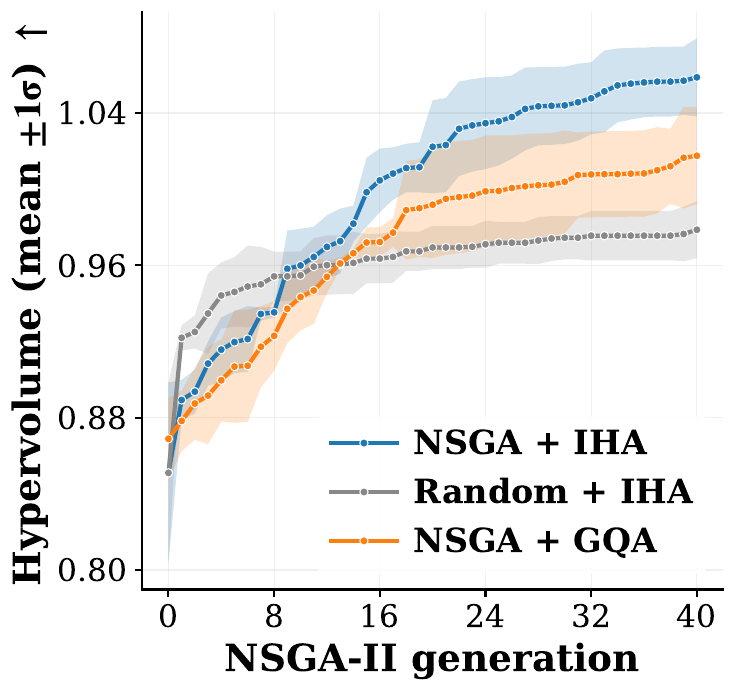}
\caption{Search-recipe ablation. Hypervolume in $(\text{val.\ loss}, \text{model size})$ vs generation, mean $\pm 1\sigma$ over seeds.}
\label{fig:ablations}
\vspace{-0.6em}
\end{wrapfigure}

\begin{figure}[t]
\centering
\includegraphics[width=\textwidth]{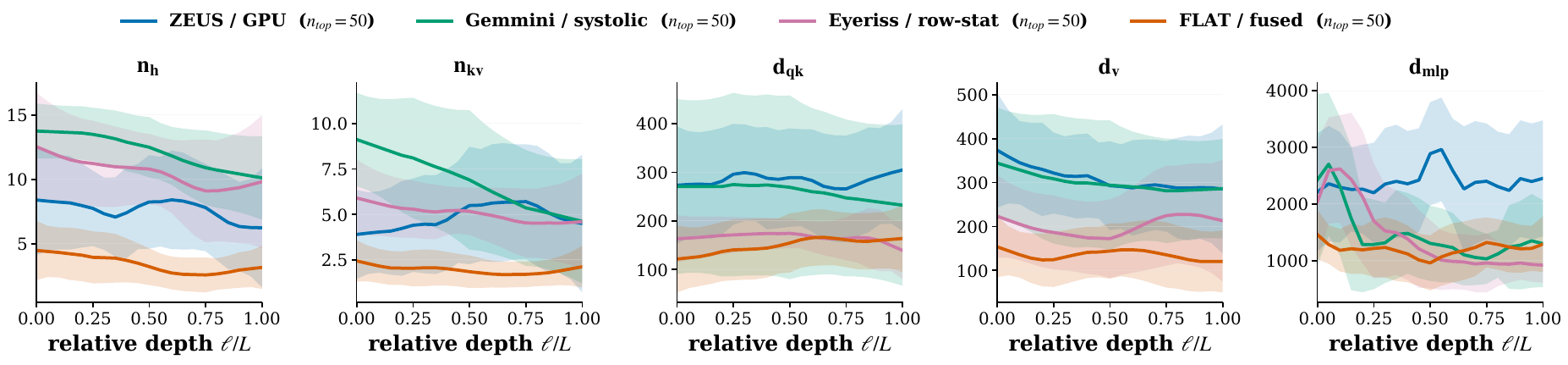}
\caption{Layer-wise architectural fingerprints of the top-50 non-dominated architectures per substrate. Solid lines are per-grid means across the 50-architecture pool, and shaded bands cover $\pm 1$ standard deviation across the pool. Active-layer field values are interpolated onto a common relative-depth grid, and layers whose attention variant is identity are masked from the attention-related fields.}
\label{fig:pareto_fingerprint}
\end{figure}

\textbf{Substrate-conditioned architectural fingerprints.} As shown in Figure~\ref{fig:pareto_fingerprint}, the four substrates pull NSGA-II toward visibly different architectures, and the divergences track each substrate's hardware bottleneck.
ZEUS on the A100 has bandwidth-rich memory and dense GEMM units, so per-layer cost is nearly uniform, and the search responds with flat MLP width and the widest attention heads of the four.
Eyeriss and Gemmini are spatial dataflows whose decode cost is dominated by KV-cache reads, which scale linearly with depth, $n_{kv}$, and $d_v$.
The search responds by collapsing $n_{kv}$ into aggressive grouped-query attention and compensating with wider MLP capacity, since MLP weights are reusable across tokens while KV must be reloaded per token.
FLAT is flexible enough to shrink every dimension evenly, producing the smallest model with shallow depth and narrow heads.
Although individual architectures across substrates can land at similar parameter counts, the depth profiles of the four substrates differ in shape.
Hardware-aware co-search therefore reshapes the per-layer width and depth profile, not just the total parameter count.

\textbf{Search-recipe ablation.} We isolate the contribution of each recipe ingredient by varying one at a time (Figure~\ref{fig:ablations}). \textbf{NSGA + IHA} is the recipe of \S\ref{subsec:main_search}, \textbf{Random + IHA} replaces NSGA-II's directed offspring with uniform sampling, and \textbf{NSGA + GQA} restricts the search to the GQA-feasible subset of IHA. NSGA + IHA reaches a final hypervolume of $1.05$ against $0.97$ for Random + IHA and $1.02$ for NSGA + GQA, isolating $0.08$ of lift from directed search and $0.03$ from IHA's expansion beyond GQA.

\subsection{rDXE Co-Search and Scaled Validation}
\label{subsec:scaled_validation}

\begin{figure}[t]
\centering
\begin{subfigure}{\textwidth}\centering
  \includegraphics[width=\textwidth]{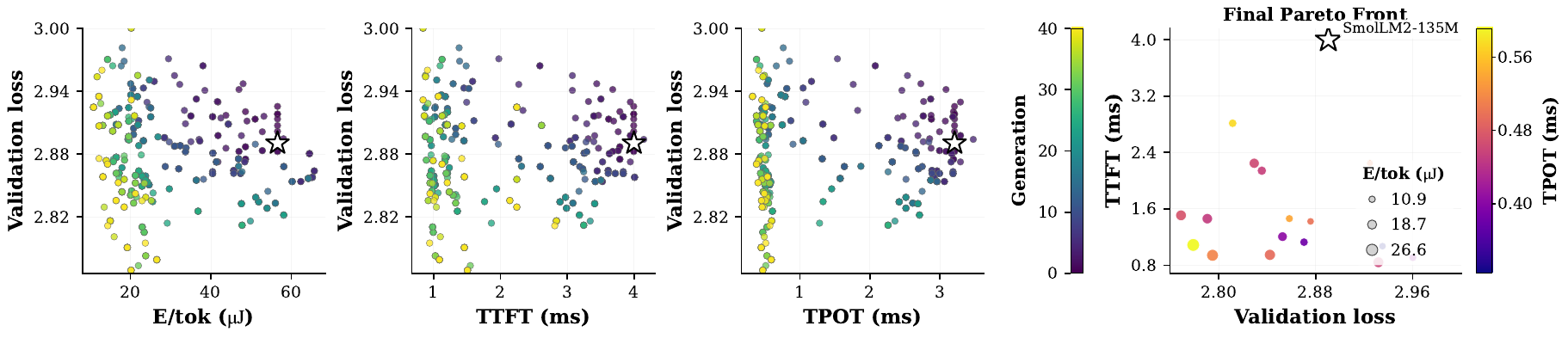}
  \caption{rDXE seeded with SmolLM2-135M.}
  \label{fig:rdxe_summary_135M}
\end{subfigure}\\[-0.3em]
\begin{subfigure}{\textwidth}\centering
  \includegraphics[width=\textwidth]{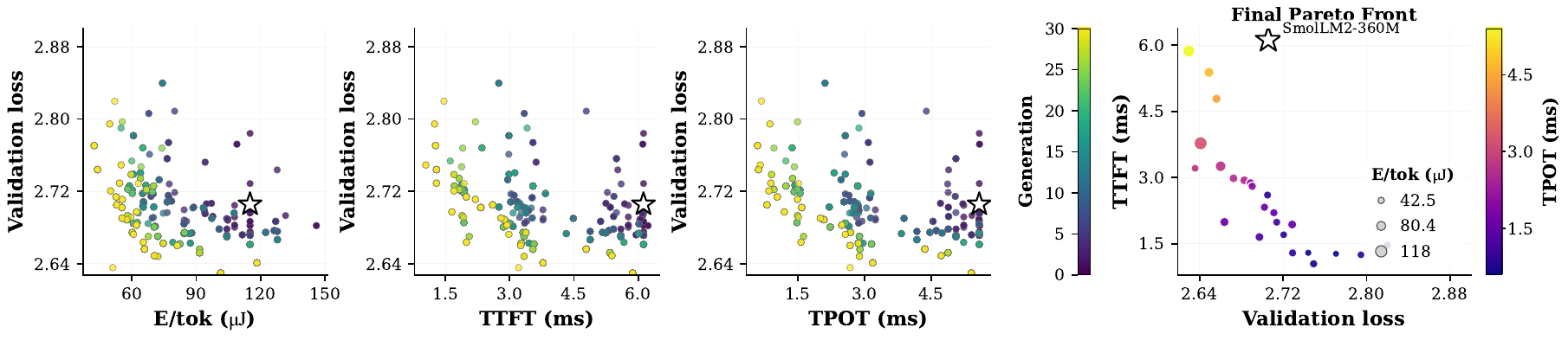}
  \caption{rDXE seeded with SmolLM2-360M.}
  \label{fig:rdxe_summary_360M}
\end{subfigure}
\caption{rDXE multi-chip ring substrate, NSGA-II evaluation with training from scratch at two seed scales. Energies are reported in $\mu$J. In the upper panel, TTFT is on the y-axis and TPOT in fill colour to surface the prefill-dominated regime at the smaller scale. Otherwise the layout matches the substrate panels of Appendix~\ref{app:cosearch_full_scatter}.}
\label{fig:rdxe_summary}
\end{figure}

Figure~\ref{fig:rdxe_summary} shows the joint architecture-and-chip Pareto fronts at both seed scales.
We retrain five Pareto-front picks, two at $\sim$100M and three at $\sim$300M, on FineWeb-Edu-10BT~\citep{penedo2024fineweb} for $13.1$B tokens under a recipe matched to architecture-matched re-trainings of SmolLM2-135M, Pythia-160M, SmolLM2-360M, and Qwen-0.5B~\citep{yang2024qwen2}, with each model evaluated on its co-searched chip configuration (Table~\ref{tab:scaled_validation}, per-pick details in Appendix~\ref{app:pareto_archs}).

\begin{table}[htbp]
\centering
\small
\caption{Scaled-training validation on FineWeb-Edu-10BT~\citep{penedo2024fineweb} ($13.1$B tokens). All models were pretrained from scratch under an identical recipe. ARC-Easy and ARC-Challenge~\citep{clark2018arc}, BoolQ~\citep{clark2019boolq}, HellaSwag~\citep{zellers2019hellaswag}, and SciQ~\citep{welbl2017sciq} are zero-shot. Hardware metrics, energy per token ($E_{\mathrm{tok}}$), time-to-first-token (TTFT), and time-per-output-token (TPOT) are evaluated on the rDXE multi-chip ring substrate at prefill 256 / decode 256. Best per metric per tier in \textbf{bold}.}
\label{tab:scaled_validation}
\setlength{\tabcolsep}{4pt}
\resizebox{\textwidth}{!}{%
\begin{tabular}{@{}llrcccccccccc@{}}
\toprule
 & & & \multicolumn{6}{c}{\textbf{SW eval}} & \multicolumn{3}{c}{\textbf{HW eval}} \\
\cmidrule(lr){4-9} \cmidrule(lr){10-12}
Tier & Model & Params (M) & val\_loss $\downarrow$ & ARC-E $\uparrow$ & ARC-C $\uparrow$ & BoolQ $\uparrow$ & HS $\uparrow$ & SciQ $\uparrow$ & $E_{\mathrm{tok}}$ ($\mu$J) $\downarrow$ & TTFT (ms) $\downarrow$ & TPOT (ms) $\downarrow$ \\
\midrule
\multirow{4}{*}{$\sim$100M}
  & SmolLM2-135M                  & 135 & 3.010          & 40.00          & \textbf{27.76} & \textbf{58.81} & 31.32          & 68.90          & 56.6          & 4.00          & 3.21 \\
  & Pythia-160M                   & 160 & 3.044          & 37.02          & 23.41          & 56.06          & 29.99          & 69.70          & 55.4          & 2.68          & 1.70 \\
  & \textbf{LLMForge-Acc-123M}    & 123 & \textbf{3.003} & \textbf{40.18} & 23.08          & 57.13          & \textbf{31.43} & 68.20          & 17.9          & 1.24          & 0.48 \\
  & \textbf{LLMForge-Compact-106M}& 106 & 3.037          & 38.95          & 23.41          & 47.65          & 30.34          & \textbf{70.70} & \textbf{14.3} & \textbf{1.13} & \textbf{0.38} \\
\midrule
\multirow{5}{*}{$\sim$300M}
  & SmolLM2-360M                  & 362 & 2.831          & \textbf{43.86} & \textbf{28.43} & \textbf{61.13} & 36.22          & 72.10          & 143.0         & 6.71          & 5.65 \\
  & Qwen-0.5B                     & 402 & 2.821          & 41.93          & 25.75          & 54.95          & 36.37          & 71.60          & 135.6         & 7.07          & 4.08 \\
  & \textbf{LLMForge-Acc-347M}    & 347 & \textbf{2.798} & 41.75          & 26.76          & 57.43          & \textbf{36.95} & \textbf{74.70} & 124.8         & 6.40          & 5.46 \\
  & \textbf{LLMForge-Eco-294M}    & 294 & 2.833          & 43.51          & \textbf{28.43} & 56.79          & 36.33          & 71.10          & \textbf{84.9} & 5.38          & 4.74 \\
  & \textbf{LLMForge-Fast-365M}   & 365 & 3.162          & 40.88          & 27.42          & 60.92          & 36.50          & 71.90          & 149.6         & \textbf{3.80} & \textbf{3.22} \\
\bottomrule
\end{tabular}}
\end{table}

In the $\sim$100M tier, LLMForge-Acc-123M tops both baselines on validation loss and HellaSwag at fewer parameters and $\sim 3\times$ lower rDXE energy per token, while LLMForge-Compact-106M trades $0.034$ in validation loss for $\sim 4\times$ lower energy and the lowest TTFT and TPOT in the tier.
In the $\sim$300M tier, LLMForge-Acc-347M reaches the lowest validation loss and highest HellaSwag and SciQ at fewer parameters than SmolLM2-360M and Qwen-0.5B, with energy per token $13\%$ below SmolLM2-360M.
LLMForge-Eco-294M and LLMForge-Fast-365M trade validation-loss headroom for $40\%$ lower energy and $43\%$ lower TTFT and TPOT respectively against the same anchor.
SmolLM2 baselines retain leads on ARC-Challenge and BoolQ at both scales, the multi-choice question-answering cluster where canonical baselines tend to hold their ground.
Together, the LLMForge picks form a deployment-aware set of architecture-and-chip configuration pairs rather than a single best model.

%% file: sections/5-conclusion.tex
\section{Conclusion}
\label{sec:conclusion}

We presented LLMForge, a hardware-aware NAS framework for edge Transformer language models that composes three contributions into a single NSGA-II loop.
Infinite-Head Attention (IHA) treats head count, KV groups, and query/key and value dimensions as independent per-layer variables, broadening the per-layer attention design space by $\sim\!400\times$ over grouped-query attention under the ranges of Table~\ref{tab:search_space}.
Forge-Former is an encoder-based accuracy surrogate that ranks IHA candidates with notably higher rank-correlation fidelity than flat MLP and tree-based baselines.
Forge-DSE plugs Forge-Former into a multi-backend cost model spanning a ZEUS-measured A100 GPU, Timeloop-modeled Gemmini, Eyeriss, and FLAT systolic accelerators, and a multi-chip ring-dataflow rDXE accelerator~\citep{tao2026rdxe}, with optional surrogate co-evolution during search.

On GPU, Gemmini, Eyeriss, and FLAT substrates, Forge-DSE converges to distinct Pareto-front architectures that consistently outperform the SmolLM2-135M baseline, with Pareto-front shapes that reflect each substrate's specific cost bottleneck.
On the multi-chip ring substrate, our co-search identifies three deployment-aware 300M-scale Pareto variants, each retrained on FineWeb-Edu-10BT under a matched recipe against SmolLM2-360M and Qwen-0.5B architecture baselines: an accuracy-oriented variant with the lowest validation loss of $2.798$ and competitive benchmark performance using fewer parameters, an energy-optimized variant that reduces energy per token by $40\%$, and a latency-optimized variant that reduces TTFT and TPOT by $43\%$.

\textbf{Limitations.}
Forge-Former is trained on the IHA design space of Appendix~\ref{app:search_space} and reliably ranks only architectures within that support. Predictions for designs outside the training corpus, such as alternative attention primitives or shape ranges beyond Table~\ref{tab:search_space}, would require retraining the surrogate on a matching labeled set.
All searches reported here target sub-billion-parameter architectures at $\le 500$M, where per-architecture pretraining is tractable within our compute budget. Extending the recipe to billion-parameter scale is the natural next step, with the main hurdle being the cost of producing a Forge-Former training corpus at that scale.

%% file: sections/7-appendix.tex
\appendix
\section*{Appendix}

\section{Search Space Specification}
\label{app:search_space}

Table~\ref{tab:search_space} lists the global and per-layer fields of the IHA-parameterized search space referenced in \S\ref{subsec:iha}.
The example domain values shown are those used by the searches reported in \S\ref{sec:results}.

\begin{table}[h]
\centering
\caption{LLMForge design space example under IHA. Global fields are shared across layers; per-layer fields are independent.}
\label{tab:search_space}
\begin{tabular}{@{}llll@{}}
\toprule
\textbf{Scope} & \textbf{Description} & \textbf{Symbol} & \textbf{Domain} \\
\midrule
\multirow{3}{*}{Global}
  & Embedding dimension & $d_{\text{model}}$ & 768 \\
  & Block size & $T$ & 1024 \\
  & Max number of layers & $L$ & 40 \\
\midrule
\multirow{7}{*}{Layer}
  & \makecell[l]{Layer gating\\\footnotesize\underline{0: pruned, 1: active}} & $\text{mask}^\ell$ & $\{0,1\}$ \\
  & \makecell[l]{Attention gating\\\footnotesize\underline{0: identity, 1: active}} & $\text{attn}^\ell$ & $\{0,1\}$ \\
  & Number of query heads & $n_h^\ell$ & $[1,16]$ \\
  & Number of KV groups & $n_{kv}^\ell$ & $[1,16]$ \\
  & Query/key dimension & $d_{qk}^\ell$ & $[64\!:\!32\!:\!512]^{\dagger}$ \\
  & Value dimension & $d_v^\ell$ & $[64\!:\!32\!:\!512]^{\dagger}$ \\
  & MLP width & $d_{mlp}^\ell$ & $[512\!:\!256\!:\!4096]^{\dagger}$ \\
\addlinespace[2pt]
\multicolumn{4}{l}{\footnotesize $^{\dagger}$Ranges $[a\!:\!\Delta\!:\!b]$ denote values from $a$ to $b$ in steps of $\Delta$.} \\
\bottomrule
\end{tabular}
\end{table}

\textbf{Configuration-count expansion.}
\label{app:iha_formal}
For $d_{\text{model}}{=}768$ and $n_h\in[1,16]$, GQA imposes $n_h\,|\,d_{\text{model}}$, $n_{kv}\,|\,n_h$, and $d_{qk}{=}d_v{=}d_{\text{model}}/n_h$, admitting $27$ configurations.
IHA replaces the divisibility and coupling constraints with independent $d_{qk}, d_v \in \{64, 96, \ldots, 512\}$, admitting $50 \times 15 \times 15 = 11{,}250$ configurations, the $\sim\!400\times$ per-layer expansion cited in \S\ref{subsec:iha}.

\section{Software Evaluation Protocol}
\label{app:sw_eval}

Each architecture used to generate ground-truth validation loss labels for Forge-Former training, and each architecture real-trained during a Forge-DSE surrogate-refinement event or post-hoc Pareto-front validation, is pretrained from scratch on MiniPile~\citep{kaddour2023minipile}, a 1.6B-token subset of The Pile~\citep{gao2020pile}, under a fixed token budget of 655M tokens on a single H100 GPU with AdamW, learning rate $1\!\times\!10^{-3}$, 100-step warmup, sequence length 512, batch size 128, and bfloat16 mixed precision.
The token budget is held fixed across all architectures so that validation loss differences reflect architectural choices rather than training-time variation.
The budget is below the Chinchilla compute-optimal point~\citep{hoffmann2022chinchilla} and is chosen to keep per-architecture training tractable at the scale of Forge-Former's training corpus.
We report validation loss at the end of this fixed budget.

\section{Forge-Former Surrogate}
\label{app:forge_former}

\subsection{Architecture and training}
\label{app:forge_former_train}

\textbf{Architecture.}
Forge-Former is a small Transformer encoder over the packed sequence of an architecture's active layers.
Each active layer is represented by a 9-dimensional feature vector consisting of the seven per-layer fields $(n_h, n_{kv}, d_{qk}, d_v, d_{mlp}, \text{mask}, \text{attn})$ and two global fields $(d_{\text{model}}, T)$ broadcast across positions, lifted to the encoder hidden dimension by a per-field linear projection $\phi_F: \mathbb{R} \to \mathbb{R}^{d_{\mathrm{enc}}}$, $F \in \{1, \dots, 9\}$.
The token at position $p$ is $z_p = \mathrm{PE}(p) + \sum_{F=1}^{9} \phi_F(x_{p,F})$, where $\mathrm{PE}$ is a learned absolute positional embedding indexed by the packed position $p \in \{0, \dots, L_{\max}-1\}$ with $L_{\max} = 40$, the IHA layer-mask cardinality.
We do not use any relative or sinusoidal positional encoding.

The encoder is a stack of $L_{\mathrm{enc}} = 4$ pre-LayerNorm blocks of width $d_{\mathrm{enc}} = 64$, with $h_{\mathrm{enc}} = 4$ self-attention heads at per-head dimension $d_{\mathrm{enc}}/h_{\mathrm{enc}} = 16$, GELU activations, FFN expansion factor $r_{\mathrm{ffn}} = 4$ for an FFN hidden width of $256$, and dropout $p_{\mathrm{drop}} = 0.2$ applied to attention weights and FFN outputs.
Predictions are obtained by masked mean pooling over non-padding positions followed by a single linear head $\mathbb{R}^{d_{\mathrm{enc}}} \to \mathbb{R}$, yielding a scalar $\hat{y}(x)$ interpreted as predicted validation loss.
A 1-layer per-field lift rather than a deeper MLP suffices for two reasons.
The inputs are normalized scalars in $[0,1]$ for which a single linear projection is sufficient, and the encoder's first attention block already mixes fields across the per-position embedding sum.
The full model has $203{,}713$ parameters.

\textbf{Training corpus.}
Forge-Former is trained on $N_{\mathrm{train}} = 1{,}642$ IHA-parameterized architectures drawn uniformly at random from the search space defined in Table~\ref{tab:search_space}, each labeled with validation loss produced under the pretraining protocol of Appendix~\ref{app:sw_eval}.
We use an $80/20$ train/test split at the architecture level, with $N_{\mathrm{test}} = 411$ and $N_{\mathrm{total}} = 2{,}053$ after dropping rows with non-finite validation loss.
The test split is held out for the ranking and regression evaluation reported in \S\ref{subsec:forge_former_ablation}.
Per-field min/max normalization statistics are computed on the training split only and broadcast to test.

\textbf{Optimization.}
We minimize the L1 loss $\mathcal{L}_1(\hat{y}, y) = |\hat{y} - y|$, chosen over MSE for its robustness to heavy-tailed residuals near the feasibility boundary of the IHA search space, using AdamW with learning rate $\eta = 1 \times 10^{-4}$, weight decay $0$, and batch size $32$.
We run for $200$ epochs with seed $100$ and report the best test-L1 checkpoint.

\subsection{Co-evolution protocol}
\label{app:co_evolution}

The co-evolving variant of Forge-Former periodically refines the surrogate during NSGA-II by real-training informative samples drawn from the live population and refitting the surrogate against the resulting labels.
This subsection details the cadence, acquisition function, fine-tuning recipe, and observed wall-clock overhead.

\textbf{Refinement cadence and acquisition budget.}
The surrogate is refit every $K = 5$ NSGA generations.
At each refinement event, $n = 8$ unevaluated architectures are dispatched to the cluster trainer, and each is trained from scratch on MiniPile under the protocol of Appendix~\ref{app:sw_eval}.
Per-architecture wall time is between $1.5$ and $2$ hours on a single H100, with the eight architectures fanned out across an 8-host pool in parallel.
A 40-generation run thus produces $8 \times 8 = 64$ acquisitions in total.

\textbf{Uncertainty estimate.}
At each refinement event, predictive uncertainty $\sigma(x)$ is obtained from $n_{\mathrm{mc}} = 10$ MC-dropout forward passes through the surrogate, with encoder dropout left active during inference.
The same $n_{\mathrm{mc}}$ samples produce the predictive mean $\mu(x)$.
Outside the acquisition step the surrogate runs in deterministic evaluation mode.

\textbf{Exploit/explore acquisition.}
Each event's batch of $n = 8$ is split half-and-half between exploitation and exploration.
The four exploit picks are the architectures with the lowest $\mu(x)$ on the first non-dominated front of the current population, and the four explore picks are the architectures with the highest $\sigma(x)$ on the remainder of the population.

\textbf{Fine-tuning recipe.}
Each refinement event re-loads the original frozen Forge-Former checkpoint $\hat{y}_0$ rather than sequentially fine-tuning from prior refits, so that later refits do not compound earlier label noise.
Fine-tuning runs for $10$ epochs with AdamW at learning rate $1 \times 10^{-4}$ and batch size $32$.
Each mini-batch interleaves samples at the replay ratio $\rho = 5.0$, drawing five rows from the $2{,}053$-row Forge-Former training corpus per one row from the cumulative real-trained buffer.
The buffer grows from $8$ architectures at event $1$ to $64$ at event $8$.
The refitted surrogate is hot-swapped into the live evaluator at the start of the next NSGA generation.

\section{Full Search Scatter Plots (Surrogate Co-Evolving)}
\label{app:cosearch_full_scatter}

\begin{figure}[h]
\centering
\begin{subfigure}{\textwidth}\centering
  \includegraphics[width=\textwidth]{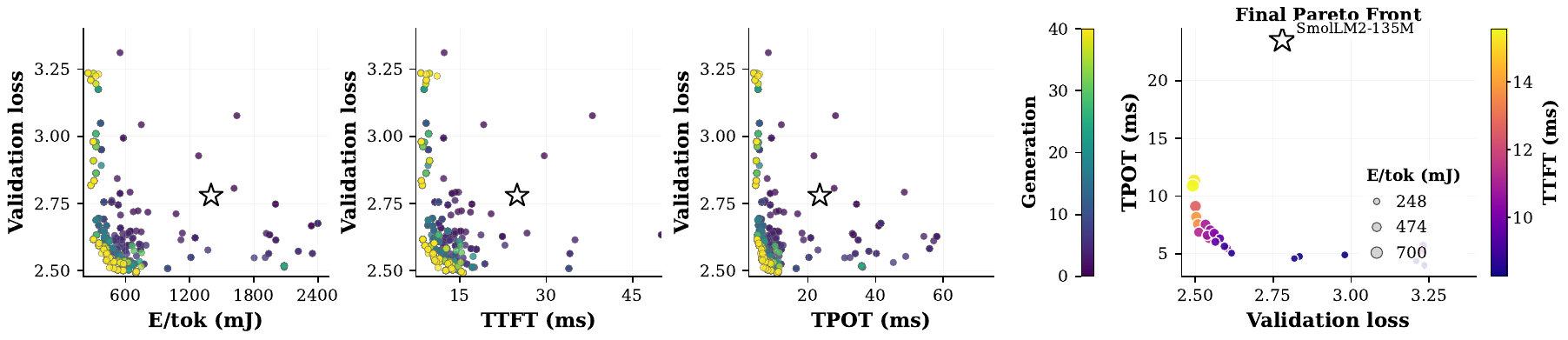}
  \caption{ZEUS A100 GPU.}
  \label{fig:zeus_summary}
\end{subfigure}\\[-0.3em]
\begin{subfigure}{\textwidth}\centering
  \includegraphics[width=\textwidth]{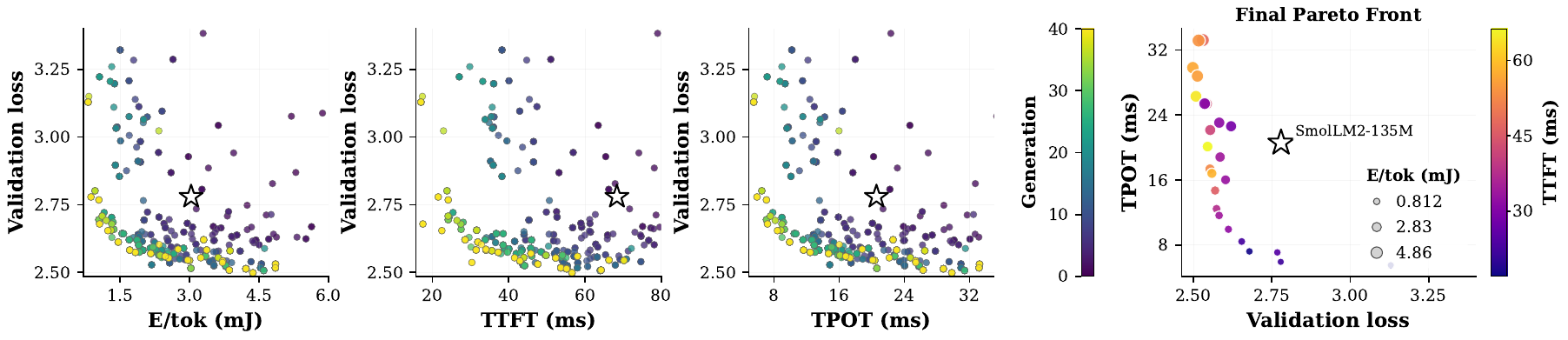}
  \caption{Gemmini systolic array (Timeloop).}
  \label{fig:gemmini_summary}
\end{subfigure}\\[-0.3em]
\begin{subfigure}{\textwidth}\centering
  \includegraphics[width=\textwidth]{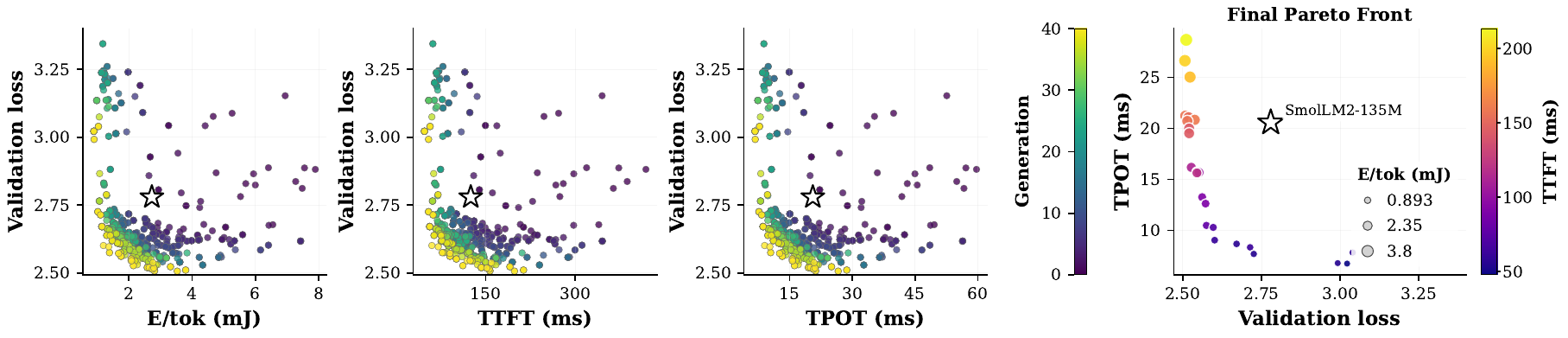}
  \caption{Eyeriss (Timeloop).}
  \label{fig:eyeriss_summary}
\end{subfigure}\\[-0.3em]
\begin{subfigure}{\textwidth}\centering
  \includegraphics[width=\textwidth]{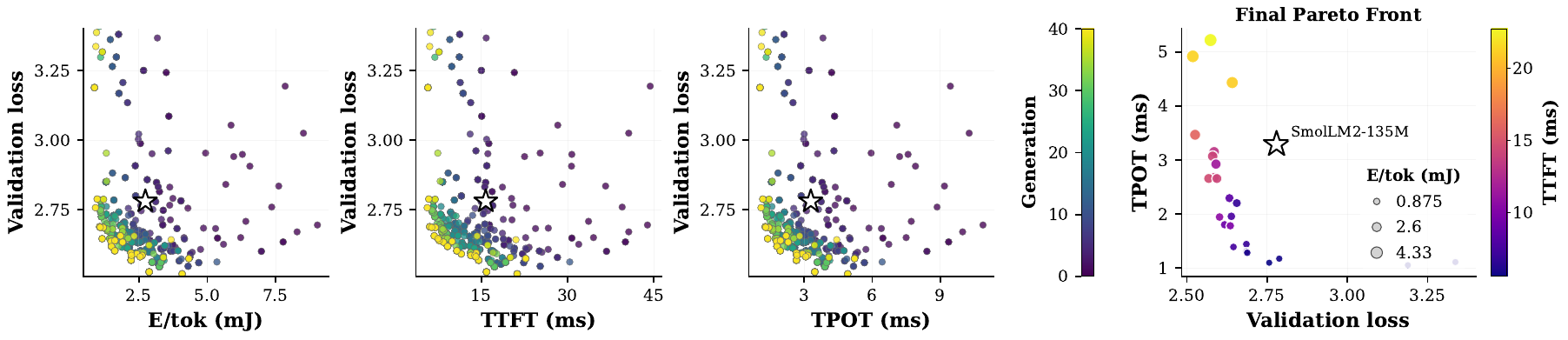}
  \caption{FLAT (Timeloop).}
  \label{fig:flat_summary}
\end{subfigure}
\caption{Full search scatter plots for the four Forge-Former + co-evolution runs. Each row shows three pairwise NSGA-II projections of validation loss vs.\ $E_{\mathrm{tok}}$, TTFT, and TPOT (coloured by generation), and the final-generation Pareto front (rightmost panel) with TTFT in fill colour and $E_{\mathrm{tok}}$ in marker size.}
\label{fig:cosearch_full_scatter}
\end{figure}

Figure~\ref{fig:cosearch_full_scatter} expands the per-substrate Pareto-front summary of \S\ref{subsec:main_search} into the full search scatter plots, showing every architecture evaluated by Forge-DSE under the surrogate-co-evolving recipe of \S\ref{subsec:forge_dse:loop}.
For each of the four hardware substrates, the figure displays three pairwise NSGA-II projections of validation loss against $E_{\mathrm{tok}}$, TTFT, and TPOT, coloured by generation index, alongside the final-generation Pareto front.
The trajectory from early-generation populations (dark) to converged Pareto fronts (light) shows where the surrogate's exploration concentrated each search and how the co-evolution refinement events shifted the population toward the substrate-specific deployment region.

\section{Hardware Backend Details}
\label{app:hw_backends}

\subsection{ZEUS GPU measurement}
\label{app:hw_zeus}

\textbf{Setup.}
All ZEUS measurements run on a single NVIDIA A100-SXM4-40GB with driver $550.90.07$ and CUDA $12.4$, paired with PyTorch $2.6.0$ and ZEUS $0.15.1$.
The energy counter source is NVML's \texttt{nvmlDeviceGetTotalEnergyConsumption}, sampled at the start and end of each measurement window via \texttt{ZeusMonitor.begin\_window} and \texttt{end\_window}.
The monitor is constructed with \texttt{sync\_execution\_with="torch"} and \texttt{approx\_instant\_energy=True}.
NVML updates this counter at a $\sim$10\,ms cadence, so when a window is shorter than the update period the monitor falls back to instantaneous-power $\times$ wall-clock-time integration rather than reporting zero energy.
Workload $\mathcal{W}$ is fixed across all GPU-substrate measurements at prefill length $L_p = 256$ followed by decode length $L_d = 256$ tokens in \texttt{bf16}, using a production-style KV-cached single-query SDPA decode loop rather than the stock recomputing \texttt{generate()} call.
Each architecture is built with random weights, warmed up for one full prefill$+$decode pass, and then measured over $n = 3$ repetitions.
We report the per-repetition median as the headline value used in NSGA-II.

\textbf{Standard error on the SmolLM2-360M anchor.}
To bound the per-architecture noise we additionally execute $N = 10$ independent measurements of the SmolLM2-360M anchor, a $362$M-parameter $32$-layer architecture at $d_{\text{model}} = 960$, under the same workload with each run consisting of one warmup pass and one timed window.
Table~\ref{tab:zeus_anchor} reports the per-metric mean, standard error of the mean, and relative standard error.
Per-token metrics are reproducible to under $0.3\%$ relative standard error, while TTFT carries the largest jitter at $\sim$3\%, dominated by CUDA-graph capture and NVML synchronization overhead at the start of the prefill window.

\begin{table}[h]
\centering
\caption{ZEUS measurement variability on the SmolLM2-360M anchor over $N{=}10$ independent runs.}
\label{tab:zeus_anchor}
\begin{tabular}{@{}lccc@{}}
\toprule
\textbf{Metric} & \textbf{Mean} & \textbf{Std.\ err.} & \textbf{Rel.\ SE} \\
\midrule
TTFT               & $28.72$\,ms  & $\pm 0.88$\,ms  & $3.05\%$ \\
TPOT               & $24.18$\,ms  & $\pm 0.05$\,ms  & $0.19\%$ \\
Power              & $62.0$\,W    & $\pm 0.09$\,W   & $0.14\%$ \\
$E_{\mathrm{tok}}$ & $1.500$\,mJ  & $\pm 0.004$\,mJ & $0.25\%$ \\
\bottomrule
\end{tabular}
\end{table}

\subsection{Timeloop-based backend}
\label{app:hw_timeloop}

Backend B in \S\ref{subsec:forge_dse:backends} instantiates four hardware templates through Timeloop's analytical cost model: Gemmini, Eyeriss, FLAT, and the single-chip DXE building block of the multi-chip rDXE backend of Appendix~\ref{app:ring_substrate}.
All four templates assume the same $16$\,nm technology library and an LPDDR4 off-chip DRAM, and differ in PE-array dimensions, on-chip SRAM hierarchy, and dataflow style.
Table~\ref{tab:timeloop_substrates} summarizes their headline parameters.

\begin{table}[h]
\centering
\caption{Timeloop-based hardware templates used as Backend B. Sizes are read from the substrate YAML configs in \texttt{nsga\_search/hw\_eval/arch/}. ``MAC precision'' lists multiplier and accumulator widths.}
\label{tab:timeloop_substrates}
\small
\begin{tabular}{@{}lcccc@{}}
\toprule
                     & \textbf{Gemmini}   & \textbf{Eyeriss}   & \textbf{FLAT}      & \textbf{Single-chip DXE} \\
\midrule
PE array             & $16{\times}16$     & $14{\times}12$     & $32{\times}32$     & $8{\times}16{\times}16$ \\
MAC count            & $256$              & $168$              & $1024$             & $2048$ \\
MAC precision (b)    & $8 / 32$           & $8 / 16$           & $8 / 16$           & $8 / 16$ \\
On-chip SRAM         & $768$\,KB          & $\sim\!200$\,KB    & $200$\,KB          & $4$\,MB \\
Dataflow             & weight-stationary  & row-stationary     & flexible           & weight-stationary \\
\bottomrule
\end{tabular}
\end{table}

\textbf{Gemmini.}
A $16 \times 16$ INT8 systolic array with a $512$\,KB scratchpad holding inputs and weights, a $256$\,KB output accumulator, and per-PE single-entry registers.
The mapper is constrained to weight-stationary or output-stationary execution.

\textbf{Eyeriss.}
A $14 \times 12$ PE array implementing the row-stationary dataflow.
Each PE owns three register files for inputs, weights, and partial sums of $24$, $384$, and $32$ bytes respectively, and a $128$\,KB shared global buffer keeps inputs and outputs while weights bypass directly to the PEs.

\textbf{FLAT.}
A $32 \times 32$ PE array with a single $200$\,KB global buffer that simultaneously keeps inputs, weights, and outputs.
This shared-residency layout is what enables on-chip fusion of attention's $QK$ and $PV$ matmuls.
The mapper is left flexible across weight-, input-, and output-stationary mappings.

\textbf{Single-chip DXE.}
The per-chip building block of the multi-chip rDXE backend in Appendix~\ref{app:ring_substrate}, not a standalone substrate.
The compute hierarchy is $8$ DXT tiles, each holding $16$ cores with $16$ INT8 multipliers, for a total of $2048$ MACs.
On-chip storage totals $4$\,MB, split as $24$\,KB of weight memory plus $8$\,KB of KV-cache storage per core, $512$\,B of head-local input buffer per DXT, and a $512$\,B chip-level global input buffer.
The dataflow is weight-stationary with weights preloaded once at boot, so DRAM is inactive during decode.

\subsection{rDXE multi-chip ring substrate}
\label{app:ring_substrate}

Backend C in \S\ref{subsec:forge_dse:backends} models a token-level pipeline of homogeneous DXE chips connected in a ring, each chip an instance of the single-chip DXE template of Appendix~\ref{app:hw_timeloop}.
For each candidate model, Backend C exhaustively explores a chip-template parameter grid, partitions the model layer sequence across the smallest number of chip instances that fits weights, KV cache, and scratchpad, simulates token-level pipelined inference on the ring, and returns the Pareto-non-dominated subset of $(\text{TTFT}, \text{TPOT}, E_{\mathrm{tok}}, A_{\text{total}})$ results, where $A_{\text{total}}$ is the sum of per-chip silicon areas.

\textbf{Chip-template parameter grid.}
For each candidate model, Backend C sweeps a $3 \times 3 \times 5 = 45$-configuration grid over three parameters, namely MACs per VAC core $n_{\text{MAC}} \in \{16, 32, 64\}$, weight memory per core $w_{\text{core}} \in \{24, 48, 96, 192, 384\}$\,KB, and a maximum ring depth $n_{\text{chips}}^{\max} \in \{8, 16, 32\}$ that bounds the layer-to-stage partitioner.
The remaining chip-template knobs are not swept directly.
The DXT-tile count $n_{\text{DXT}}$ per chip and the VAC core count $n_{\text{VAC}}$ per tile are derived per pick by the WMEM-packing routine described below, which factors the required core count into a power-of-2 split with $n_{\text{VAC}} \geq n_{\text{DXT}}$.
KV-cache per core $k_{\text{core}}$ is held at the chip-template default of $8$\,KB throughout this study.
Per-chip area is approximated as a linear function of total on-chip SRAM capacity, anchored to the single-chip DXE reference template.

\textbf{Per-architecture evaluation flow.}
For each candidate model, Backend C executes the following five steps.
\begin{enumerate}[itemsep=0pt, topsep=2pt, leftmargin=*]
  \item \emph{Layer profiling.}
  Each active layer $\ell$ is profiled for $(w_\ell, \kappa_\ell, o_\ell, a_\ell)$, namely weight bytes, KV bytes per token, decode ops, and peak activation bytes, via \texttt{hw\_packer.profile\_model}.
  \item \emph{Chip-config feasibility.}
  Each of the $45$ chip configs is passed to the balanced contiguous packing routine of step~3, which returns $\varnothing$ when no valid partition exists for the given chip template. Infeasible configs are skipped, leaving a viable subset for ring simulation.
  \item \emph{Balanced contiguous packing.}
  For each viable chip config, Algorithm~\ref{alg:balanced_pack} assigns layers to a minimal number of contiguous chip stages while minimizing the longest-stage decode latency.
  \item \emph{Ring simulation.}
  The resulting ring is simulated by the rDXE token-level pipelined simulator under the prefill$+$decode workload of Table~\ref{tab:nsga_config}, returning TTFT, TPOT, and energy per token.
  \item \emph{Pareto filtering.}
  The chip-config results are jointly Pareto-filtered on $(\text{TTFT}, \text{TPOT}, E_{\mathrm{tok}}, A_{\text{total}})$ and the top-$K$ non-dominated $(\text{model}, \text{chip-config})$ pairs are returned, with $K = 3$ by default.
\end{enumerate}
The retained pairs feed back into the NSGA-II Pareto selection of \S\ref{subsec:forge_dse:loop}.
A model with no feasible chip config is assigned infinite hardware metrics and is dominated out of the next generation.

\textbf{Balanced contiguous packing.}
Algorithm~\ref{alg:balanced_pack} binary-searches over the maximum allowed per-chip decode-ops budget $B$ to find the smallest budget that admits a feasible contiguous partition.
The inner feasibility test is the greedy contiguous partitioner of Algorithm~\ref{alg:greedy_partition}, which scans layers in order, extending the current stage while WMEM, KV, scratchpad, and ops budgets all hold and sealing the stage off otherwise.
The optimum $B$ minimizes the longest-stage decode time, the bottleneck of token-level pipelined throughput.

\begin{algorithm}[h]
\caption{\textsc{BalancedContiguousPack}: minimize longest-stage decode time over contiguous partitions.}
\label{alg:balanced_pack}
\begin{algorithmic}[1]
\REQUIRE Layer costs $\mathcal{L} = [\ell_1, \dots, \ell_N]$, chip spec $C$, max context $T$.
\ENSURE Partition $\mathcal{P}^\star$, or $\varnothing$ if infeasible.
\IF{any single $\ell_i$ violates $C$} \RETURN $\varnothing$ \ENDIF
\STATE $B_{\min} \gets \max_i \ell_i.o$;\quad $B_{\max} \gets \sum_i \ell_i.o$;\quad $\mathcal{P}^\star \gets \varnothing$
\WHILE{$B_{\min} \le B_{\max}$}
  \STATE $B \gets \lfloor (B_{\min} + B_{\max}) / 2 \rfloor$
  \STATE $\mathcal{P} \gets \textsc{GreedyContiguousPartition}(\mathcal{L}, C, T, B)$
  \IF{$\mathcal{P} \neq \varnothing$}
    \STATE $\mathcal{P}^\star \gets \mathcal{P}$;\quad $B_{\max} \gets B - 1$
  \ELSE
    \STATE $B_{\min} \gets B + 1$
  \ENDIF
\ENDWHILE
\RETURN $\mathcal{P}^\star$
\end{algorithmic}
\end{algorithm}

\begin{algorithm}[h]
\caption{\textsc{GreedyContiguousPartition}: inner feasibility check used by Algorithm~\ref{alg:balanced_pack}.}
\label{alg:greedy_partition}
\begin{algorithmic}[1]
\REQUIRE Layers $\mathcal{L}=[\ell_1,\dots,\ell_N]$ with per-layer costs $(\ell.w,\ell.\kappa,\ell.o,\ell.a)$ for weight bytes, KV bytes/token, decode ops, and peak activation bytes; chip $C=(W,K,A,T)$ for weight mem, KV cache, scratchpad, and max context; stage budget $B$.
\ENSURE Contiguous partition $\mathcal{P}$, or $\varnothing$ if infeasible.
\STATE $\mathcal{P}\gets[\,]$;\quad $\mathcal{S}\gets[\,]$;\quad $(w,k,o)\gets(0,0,0)$
\FOR{$i=1$ to $N$}
    \STATE $\ell\gets\mathcal{L}[i]$;\quad $\Delta k\gets \ell.\kappa\cdot T$
    \IF{$\ell.w>W$ \textbf{or} $\Delta k>K$ \textbf{or} $\ell.o>B$ \textbf{or} $\ell.a>A$}
        \RETURN $\varnothing$ \quad\textit{// layer alone exceeds chip}
    \ENDIF
    \IF{$w+\ell.w\le W$ \textbf{and} $k+\Delta k\le K$ \textbf{and} $o+\ell.o\le B$}
        \STATE append $\ell$ to $\mathcal{S}$;\quad $(w,k,o)\gets(w+\ell.w,\,k+\Delta k,\,o+\ell.o)$
    \ELSE
        \STATE append $\mathcal{S}$ to $\mathcal{P}$;\quad $\mathcal{S}\gets[\ell]$;\quad $(w,k,o)\gets(\ell.w,\Delta k,\ell.o)$
    \ENDIF
\ENDFOR
\IF{$\mathcal{S}\neq[\,]$} \STATE append $\mathcal{S}$ to $\mathcal{P}$ \ENDIF
\RETURN $\mathcal{P}$
\end{algorithmic}
\end{algorithm}

\section{Forge-DSE Search}
\label{app:forge_dse}

\subsection{Algorithm pseudocode}
\label{app:forge_dse_alg}

Algorithm~\ref{alg:forge_dse} formalizes the Forge-DSE search procedure described in \S\ref{subsec:forge_dse:loop}.
The pseudocode covers the full NSGA-II generation loop, the mutation--crossover--repair pipeline applied to each offspring, and the optional surrogate-refinement event that fine-tunes Forge-Former when $K > 0$.

\begin{algorithm}[h]
\small
\caption{Forge-DSE: NSGA-II search with optional co-evolving Forge-Former surrogate ($K{=}0$ disables refinement).}
\label{alg:forge_dse}
\begin{algorithmic}[1]
\REQUIRE Search space $\mathcal{A}$, objective set $F$, constraint set $B$, Forge-Former $\hat{y}(\cdot)$ with frozen baseline $\hat{y}_0$, HW backend $\mathrm{HW}(\cdot, \mathcal{W})$ with fixed workload $\mathcal{W}$, population size $N$, offspring size $\lambda$, generations $G$, crossover rate $p_c$, mutation rate $p_m$, refinement cadence $K$ ($K{=}0$ disables co-evolution), refinement batch $b$, MC-dropout passes $n_{\mathrm{mc}}$, original training corpus $\mathcal{D}_0$ with old:new replay ratio $\rho$
\ENSURE Pareto archive $\mathcal{P}$ of all non-dominated feasible architectures encountered during search
\STATE Initialize $P^{(0)} \subset \mathcal{A}$, $|P^{(0)}| = N$; evaluate $F$ and $B$ for all $x \in P^{(0)}$ using $\hat{y}(x)$ and $\mathrm{HW}(x, \mathcal{W})$
\STATE $\mathcal{P} \gets \emptyset$;\quad $\mathcal{D} \gets \emptyset$ \quad // Pareto archive; co-evolution label buffer (grown in place across events)
\STATE $\hat{y} \gets \hat{y}_0$ \quad // mutable working surrogate; $\hat{y}_0$ stays frozen as the refit anchor
\FOR{$t = 0$ to $G-1$}
    \STATE $M^{(t)} \gets \textsc{TournamentSelect}(P^{(t)})$
    \STATE $Q^{(t)} \gets \emptyset$
    \WHILE{$|Q^{(t)}| < \lambda$}
        \STATE Sample $p_1, p_2 \in M^{(t)}$
        \STATE $c \gets \textsc{Crossover}(p_1, p_2)$ w.p.\ $p_c$, else $c \gets p_1$
        \STATE $c \gets \textsc{Mutate}(c)$ w.p.\ $p_m$
        \STATE $Q^{(t)} \gets Q^{(t)} \cup \{\textsc{Repair}(c)\}$
    \ENDWHILE
    \STATE Evaluate $F$ and $B$ for $x \in Q^{(t)}$ using $\hat{y}(x)$ and $\mathrm{HW}(x, \mathcal{W})$ \quad // $\hat{y}$ reflects any refinement applied in prior events
    \STATE $R^{(t)} \gets P^{(t)} \cup Q^{(t)}$
    \STATE $P^{(t+1)} \gets \textsc{Survival}(R^{(t)}, N)$ \quad // NSGA-II constraint-dominance + crowding distance~\citep{nsgaII}
    \STATE $\mathcal{P} \gets \textsc{UpdateParetoSet}(\mathcal{P}, R^{(t)}_{\text{feas}})$ \quad // $R^{(t)}_{\text{feas}}$: feasible subset of $R^{(t)}$ under $B$
    \IF{$K > 0$ \textbf{and} $t > 0$ \textbf{and} $t \bmod K = 0$}
        \STATE \textit{// surrogate-refinement event (skip $t{=}0$: no offspring yet)}
        \STATE $(\mu, \sigma) \gets \textsc{MCDropout}(\hat{y}, P^{(t+1)}, n_{\mathrm{mc}})$ \quad // per-$x$ arrays from $n_{\mathrm{mc}}$ stochastic forwards
        \STATE $\mathcal{F} \gets \textsc{FastNonDominatedSort}(P^{(t+1)})[0]$ \quad // first front under $(F, B)$
        \STATE $b_{\text{exp}} \gets \min(\lfloor b/2 \rfloor, |\mathcal{F}|)$ \quad // exploitation quota, capped by front size
        \STATE $E_{\text{exploit}} \gets$ top $b_{\text{exp}}$ of $\mathcal{F}$ by ascending $\mu$ \quad // lowest predicted validation loss
        \STATE $E_{\text{explore}} \gets$ top $\min(b - b_{\text{exp}},\, |P^{(t+1)} \setminus \mathcal{F}|)$ of $P^{(t+1)} \setminus \mathcal{F}$ by descending $\sigma$
        \STATE $E \gets E_{\text{exploit}} \cup E_{\text{explore}}$
        \STATE $\mathcal{L} \gets \{(x, \textsc{RealTrain}(x)) : x \in E\}$ \quad // protocol of Appendix~\ref{app:sw_eval}
        \STATE $\mathcal{D} \gets \mathcal{D} \cup \{(x, y) \in \mathcal{L} : y \text{ finite}\}$ \quad // drop NaN/inf labels
        \STATE $\hat{y} \gets \textsc{FineTune}(\hat{y}_0,\, \mathcal{D},\, \mathcal{D}_0,\, \rho)$ \quad // refit on $\mathcal{D} \cup \mathcal{D}_0$ sampled at $\rho{:}1$ old:new
    \ENDIF
\ENDFOR
\RETURN $\mathcal{P}$
\end{algorithmic}
\end{algorithm}

\subsection{Per-experiment NSGA-II configurations}
\label{app:nsga_config}

Table~\ref{tab:nsga_config} reports the NSGA-II settings for each of the five searches presented in \S\ref{sec:results}.
The four Forge-Former driven searches (ZEUS, Gemmini, Eyeriss, FLAT) share the same surrogate-with-co-evolution recipe and differ only in the hardware backend used to evaluate latency and energy.
The rDXE search instead trains every evaluated candidate from scratch and runs at a smaller per-evaluation budget under tighter feasibility gates that keep the population on the SmolLM2-360M scale.

\begin{table}[h]
\centering
\caption{Per-experiment NSGA-II configuration for the five searches in \S\ref{sec:results}. The four Forge-Former-driven runs share the surrogate, co-evolution cadence, and budget, and differ only in the HW backend (top rows). The rDXE search uses training-from-scratch evaluation and tighter feasibility constraints.}
\label{tab:nsga_config}
\small
\begin{tabular}{@{}lcccc@{\hspace{1.5em}}c@{}}
\toprule
                & \multicolumn{4}{c}{\textbf{Forge-Former + co-evolution}} & \textbf{Real training} \\
\cmidrule(lr){2-5} \cmidrule(lr){6-6}
\textbf{Setting} & \textbf{ZEUS}    & \textbf{Gemmini}  & \textbf{Eyeriss} & \textbf{FLAT} & \textbf{rDXE} \\
\midrule
HW backend                & ZEUS (A100)  & Timeloop  & Timeloop  & Timeloop  & rDXE simulator \\
Substrate                 & n/a  & gemmini   & eyeriss   & flat\_edge & multi-chip ring \\
Population size           & 24 & 24 & 24 & 24 & 24 \\
Offspring size            & 48 & 48 & 48 & 48 & 12 \\
Generations               & 40 & 40 & 40 & 40 & 20 \\
Crossover / mutation rate & \multicolumn{4}{c}{0.6 / 0.3} & 0.6 / 0.3 \\
Co-evolution cadence (batch) & \multicolumn{4}{c}{every 5 gens (8 archs / refit)} & n/a \\
Prefill / decode (tokens) & \multicolumn{4}{c}{256 / 256} & 512 / 256 \\
Objectives (all $\downarrow$) & \multicolumn{5}{c}{val.\ loss, TTFT, TPOT, $E_{\mathrm{tok}}$} \\
Val.\ loss constraint     & \multicolumn{4}{c}{$< 3.8$} & $< 3.5$ \\
\bottomrule
\end{tabular}
\end{table}

\section{Experimental Details}
\label{app:exp_details}

\subsection{Baseline reproduction}
\label{app:baselines}

\textbf{Baseline architectures.}
The four baselines reported in Table~\ref{tab:scaled_validation}, namely SmolLM2-135M, SmolLM2-360M, Pythia-160M, and Qwen-0.5B, are pretrained from scratch under the same recipe as the LLMForge candidates rather than loaded from upstream checkpoints.
We adopt each model's published architecture, specifically per-layer hidden width, depth, attention heads, GQA ratios, activation, norm placement, and positional encoding, but instantiate it with our shared $50{,}257$-token GPT-2 BPE tokenizer rather than the original.
Training from scratch on FineWeb-Edu sample-10BT~\citep{penedo2024fineweb} under our shared recipe removes tokenizer, training-corpus, and training-compute confounders relative to the upstream checkpoints.
The architecture references are \texttt{HuggingFaceTB/SmolLM2-135M}~\citep{allal2025smollm2}, \texttt{HuggingFaceTB/SmolLM2-360M}~\citep{allal2025smollm2}, \texttt{EleutherAI/pythia-160m}~\citep{biderman2023pythia}, and \texttt{Qwen/Qwen2.5-0.5B}~\citep{yang2024qwen2}.
Architecture YAML configs for each baseline are released alongside this paper at \texttt{fineweb\_baselines/config/standard\_baselines.yaml}.

\textbf{Training recipe.}
All entries in Table~\ref{tab:scaled_validation}, baselines and LLMForge picks alike, are trained from random initialization with identical hyperparameters.
We use AdamW with $\beta_1 = 0.9$, $\beta_2 = 0.99$, weight decay $0.1$, and $\epsilon = 10^{-8}$.
The learning rate follows a cosine schedule from $3 \times 10^{-4}$ to $3 \times 10^{-5}$ over $100{,}000$ iterations with $2{,}000$ linear warmup steps, and global $L_2$ gradient clipping is applied at $1.0$ with no dropout.
Each optimizer step processes $64$ sequences of $1{,}024$ tokens at $2$ gradient-accumulation steps, totalling $131{,}072$ tokens per step and $\approx 13.1$\,B tokens over the full schedule.
Training runs in bfloat16 mixed precision with \texttt{torch.compile}, and gradient checkpointing is enabled for all models above $100$\,M parameters.
The tokenizer is the GPT-2 BPE \texttt{tiktoken} build with $50{,}257$ vocabulary entries, and the dataset is FineWeb-Edu sample-10BT with a held-out validation slice used for the val\_loss column.
Each run executes on a single NVIDIA H100 80\,GB GPU.

\textbf{Evaluation.}
Zero-shot benchmark accuracies are computed with our custom evaluation harness using length-normalized log-likelihood scoring.
For each multiple-choice example with continuations $c_1, \ldots, c_K$ and shared context $x$, the predicted continuation is
\[
\hat{c} \;=\; \arg\max_{k}\; \frac{1}{|c_k|}\sum_{t}\log p_\theta\!\left(c_k^{(t)} \mid x,\, c_k^{(<t)}\right).
\]
All evaluations use a block size of $1{,}024$, matching training, and run in bfloat16.

\subsection{Pareto-front architecture and chip configurations}
\label{app:pareto_archs}

\begin{figure}[t]
\centering
\includegraphics[width=\textwidth]{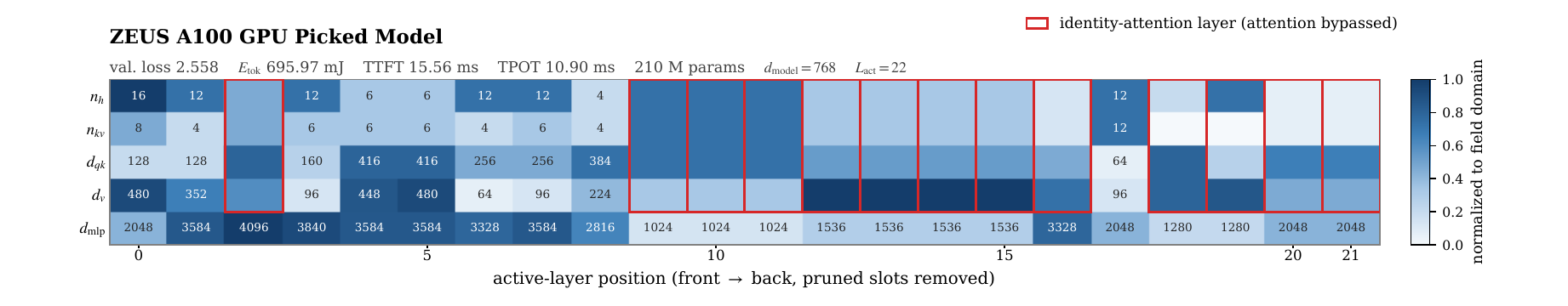}\\[0.3em]
\includegraphics[width=\textwidth]{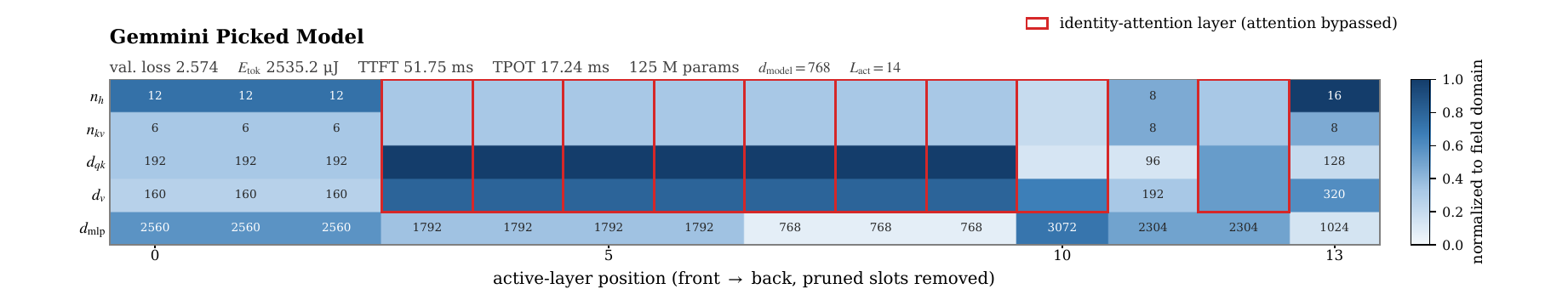}\\[0.3em]
\includegraphics[width=\textwidth]{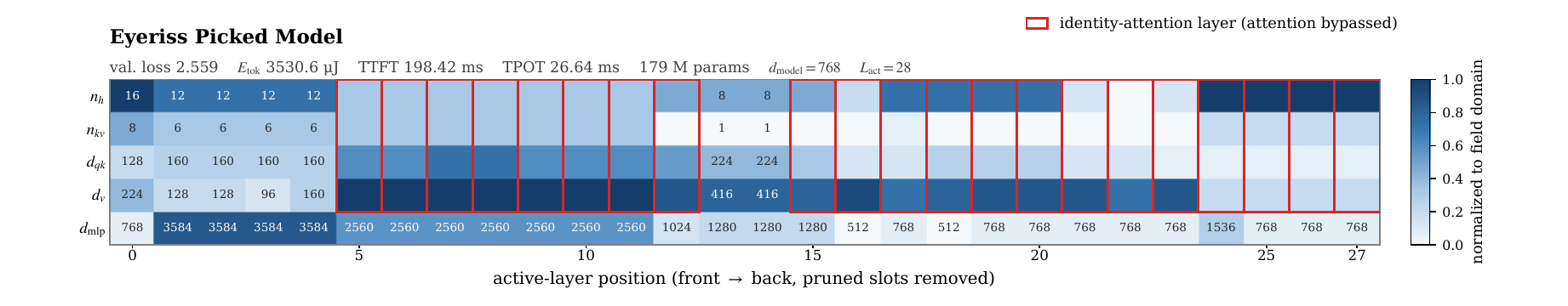}\\[0.3em]
\includegraphics[width=\textwidth]{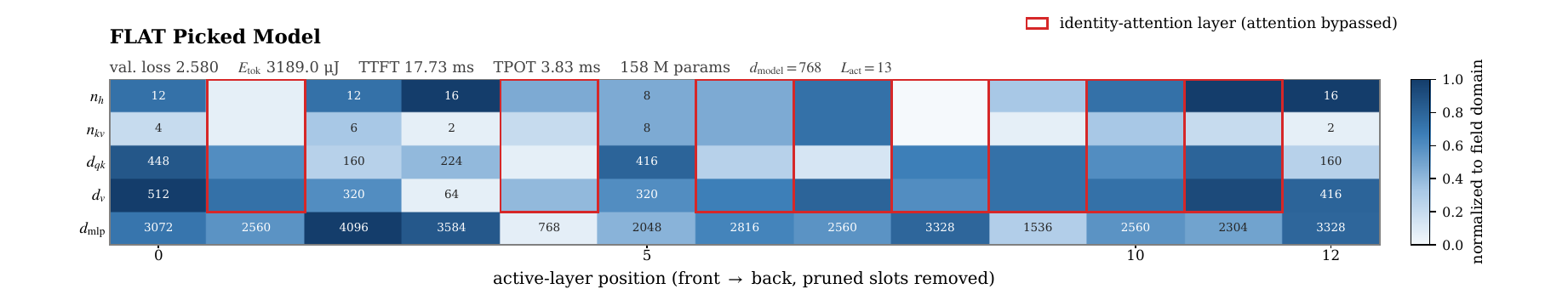}
\caption{Per-substrate Pareto-front picks corresponding to the highlighted points in Figure~\ref{fig:substrate_pareto_4row}: ZEUS A100 GPU, Gemmini, Eyeriss, and FLAT (top to bottom).}
\label{fig:appdx_arch_substrates}
\end{figure}

\begin{figure}[t]
\centering
\includegraphics[width=\textwidth]{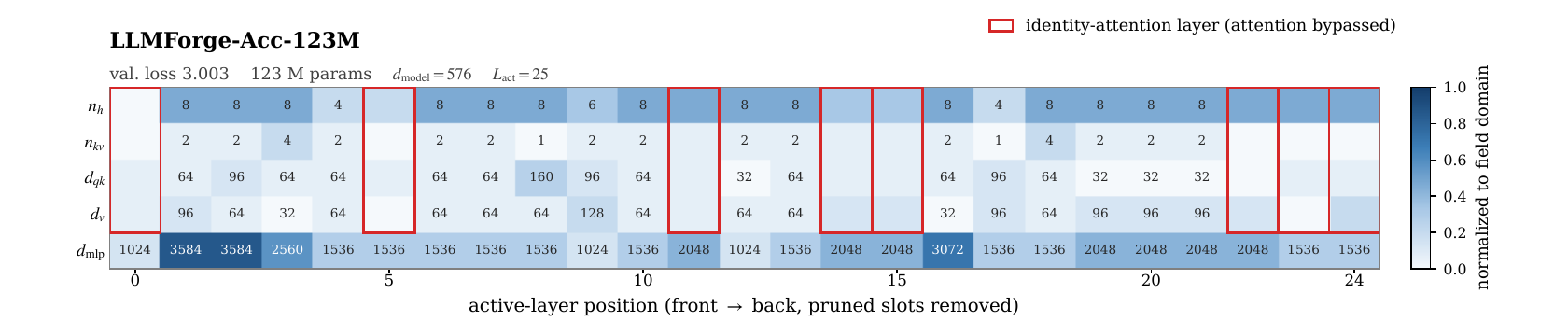}\\[0.3em]
\includegraphics[width=\textwidth]{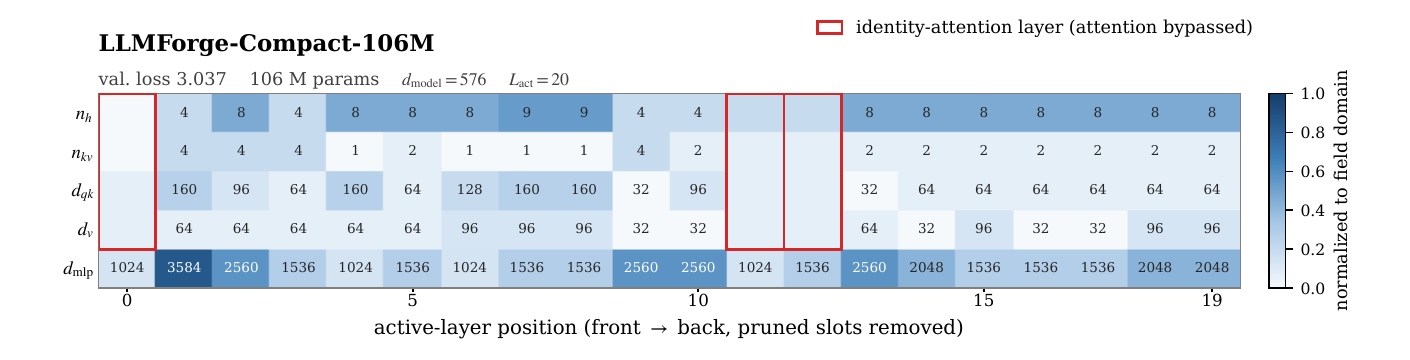}
\caption{$\sim$100M-tier picks from Table~\ref{tab:scaled_validation}: \textbf{LLMForge-Acc-123M} (top, val.\ loss leader) and \textbf{LLMForge-Compact-106M} (bottom, lowest energy per token, TTFT, and TPOT).}
\label{fig:appdx_arch_100m}
\end{figure}

\begin{figure}[t]
\centering
\includegraphics[width=\textwidth]{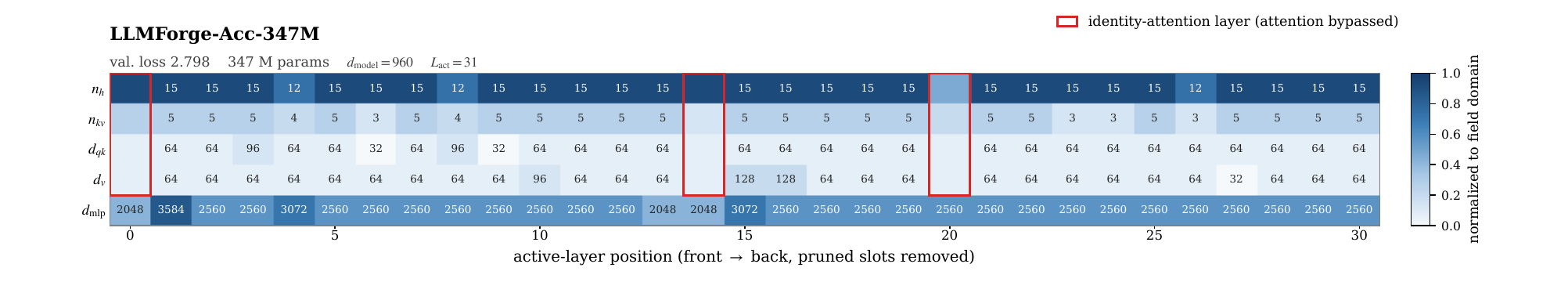}\\[0.3em]
\includegraphics[width=\textwidth]{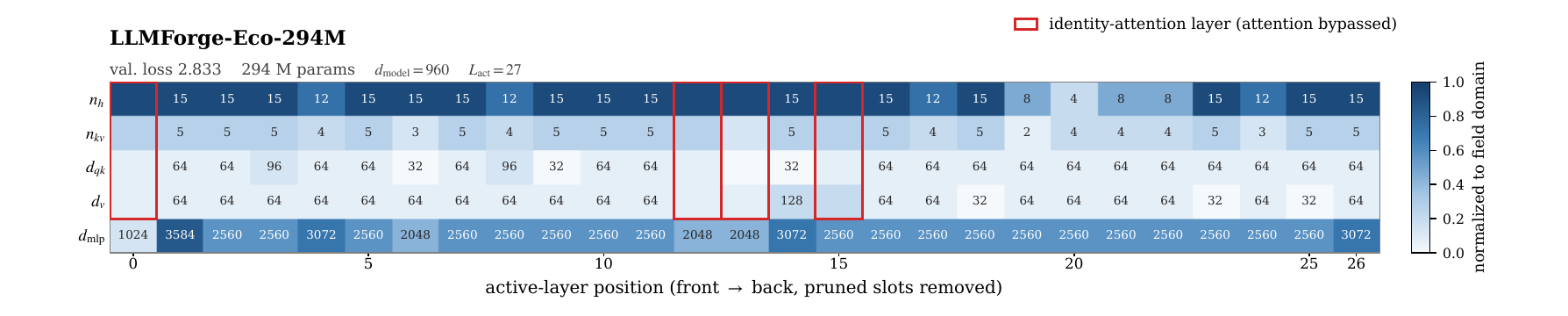}\\[0.3em]
\includegraphics[width=\textwidth]{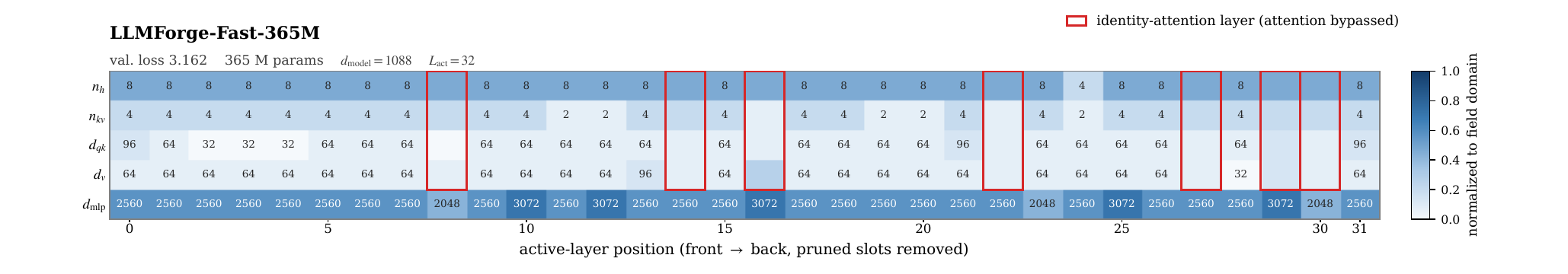}
\caption{$\sim$300M-tier picks from Table~\ref{tab:scaled_validation}: \textbf{LLMForge-Acc-347M} (top, val.\ loss leader), \textbf{LLMForge-Eco-294M} (middle, lowest energy per token), and \textbf{LLMForge-Fast-365M} (bottom, lowest TTFT and TPOT).}
\label{fig:appdx_arch_300m}
\end{figure}

Figure~\ref{fig:appdx_arch_substrates} shows per-layer IHA parameterizations of the architectures highlighted in Figure~\ref{fig:substrate_pareto_4row}, one Pareto-front pick per Forge-Former-driven substrate. The five LLMForge picks reported in Table~\ref{tab:scaled_validation} are shown in Figures~\ref{fig:appdx_arch_100m} and~\ref{fig:appdx_arch_300m}. Each panel encodes $n_h$, $n_{kv}$, $d_{qk}$, $d_v$, and $d_{\mathrm{mlp}}$ across active layers, with pruned slots removed. Cell colour shows the value normalised to its field domain, and red frames mark identity-attention layers, where the attention block is bypassed.

\begin{table}[t]
\centering
\small
\caption{Ring chip configurations selected by the rDXE chip-Pareto inner loop for every model in Table~\ref{tab:scaled_validation}. \emph{nMAC} is MACs per VAC core and \emph{wcore} is per-core weight memory, both swept directly along with the ring-depth cap $n_{\text{chips}}^{\max} \in \{8, 16, 32\}$ described in Appendix~\ref{app:ring_substrate}. \emph{n\textsubscript{chips}} is the realised ring depth, bounded above by $n_{\text{chips}}^{\max}$. \emph{nDXT}\,$\times$\,\emph{nVAC} is derived per pick by the WMEM-packing routine, factoring the required core count into a power-of-2 split with nVAC\,$\geq$\,nDXT. \emph{kcore} is held at the chip-template default of $8$~KB per core throughout this study.}
\label{tab:rdxe_chip_configs}
\setlength{\tabcolsep}{6pt}
\begin{tabular}{@{}lrrcrr@{}}
\toprule
Model & nMAC & wcore (KB) & nDXT\,$\times$\,nVAC & kcore (KB) & $n_{\text{chips}}$ \\
\midrule
SmolLM2-135M             & 16 & 24 & 16$\times$32 & 8 &  8 \\
Pythia-160M              & 16 & 24 & 32$\times$32 & 8 &  4 \\
\textbf{LLMForge-Acc-123M}     & 16 & 24 & 16$\times$16 & 8 &  7 \\
\textbf{LLMForge-Compact-106M} & 16 & 24 & 16$\times$16 & 8 &  5 \\
\midrule
SmolLM2-360M             & 16 & 48 & 32$\times$32 & 8 &  7 \\
Qwen-0.5B                & 16 & 96 & 16$\times$32 & 8 &  6 \\
\textbf{LLMForge-Acc-347M}     & 16 & 24 & 32$\times$32 & 8 & 11 \\
\textbf{LLMForge-Eco-294M}     & 16 & 24 & 32$\times$32 & 8 &  7 \\
\textbf{LLMForge-Fast-365M}    & 16 & 24 & 32$\times$64 & 8 &  7 \\
\bottomrule
\end{tabular}
\end{table}

\section{Reproducibility, Compute, and Broader Impact}
\label{app:reproducibility}

\subsection{Compute disclosure}
\label{app:compute}

All training runs in this work execute on NVIDIA H100 GPUs.

\textbf{Train-from-scratch runs.}
Each per-architecture train-from-scratch run, used for Forge-Former label generation, surrogate-refinement events, and post-hoc Pareto-front pre-validation, executes $20{,}000$ iterations on MiniPile under the protocol of Appendix~\ref{app:sw_eval} on a single H100.

\textbf{Pareto-front validation pretraining.}
Pareto-front candidates selected for the scaled validation in \S\ref{subsec:scaled_validation} are pretrained on FineWeb-Edu-10BT for $100{,}000$ iterations on a single H100 each.

\textbf{NSGA-II searches.}
Inside an NSGA-II run, the eight candidate trainings produced at each surrogate-refinement event are dispatched to an 8-host H100 pool and run in parallel.
Each refinement event therefore completes in one per-architecture wall time rather than eight.

\subsection{Datasets, code, and weights}
\label{app:release}

All datasets used in this work, namely MiniPile~\citep{kaddour2023minipile}, FineWeb-Edu-10BT~\citep{penedo2024fineweb}, HellaSwag~\citep{zellers2019hellaswag}, and ARC-E~\citep{clark2018arc}, are publicly available on HuggingFace under permissive research-use licenses.
Code, Forge-Former weights, and the per-substrate Pareto-front architecture configurations are released at the anonymized supplementary URL \url{https://anonymous.4open.science/r/llmforge_code-6838/llmforge}.

\subsection{Broader impact}
\label{app:broader_impact}

LLMForge is a hardware-aware neural architecture search framework rather than a new pretrained model.
Its contributions are methodological, namely the IHA attention parameterization, the Forge-Former surrogate, and the Forge-DSE multi-backend search engine.
The architectures it discovers at sub-billion-parameter scale yield modest energy and latency reductions on the substrates we report, which primarily benefits the operating cost and carbon profile of repeated edge inference rather than enabling new model capabilities.
We do not see any risk specific to this work beyond those already shared by small edge-deployed language models, namely downstream misuse without server-side moderation and biases inherited from public web-text pretraining corpora.